\documentclass[journal]{IEEEtran}

\usepackage[english]{babel}
\usepackage{cite}
\usepackage{lipsum}
\usepackage{amsmath,amssymb,amsfonts}
\usepackage{graphicx}
\usepackage{textcomp}
\usepackage{xcolor}
\usepackage{float}
\usepackage{subcaption}
\usepackage{algorithm}
\usepackage{algpseudocode}
\usepackage{mwe}
\usepackage[flushleft]{threeparttable}
\usepackage{color}
\usepackage{colortbl}
\usepackage{booktabs}
\usepackage{multirow}
\usepackage{url}
\usepackage{xcolor}
\newcommand{\ra}[1]{\renewcommand{\arraystretch}{#1}}

\makeatletter 
\newenvironment{breakablealgorithm}
  {
   \begin{center}
     \refstepcounter{algorithm}
     \hrule height.8pt depth0pt \kern2pt
     \renewcommand{\caption}[2][\relax]{
       {\raggedright\textbf{\ALG@name~\thealgorithm} ##2\par}%
       \ifx\relax##1\relax 
         \addcontentsline{loa}{algorithm}{\protect\numberline{\thealgorithm}##2}%
       \else 
         \addcontentsline{loa}{algorithm}{\protect\numberline{\thealgorithm}##1}%
       \fi
       \kern2pt\hrule\kern2pt
     }
  }{
     \kern2pt\hrule\relax
   \end{center}
  }
\makeatother

\newcommand{\headersty}[1]{{\normalfont\normalsize\centering\scshape #1}}
\newcommand{\unaryminus}{\scalebox{0.75}[1.0]{\( - \)}}
\newcommand{\algrule}[1][.2pt]{\par\vskip.5\baselineskip\hrule height #1\par\vskip.5\baselineskip}
  
\def\BibTeX{{\rm B\kern-.05em{\sc i\kern-.025em b}\kern-.08em
    T\kern-.1667em\lower.7ex\hbox{E}\kern-.125emX}}
    
\ifCLASSINFOpdf
\else
\fi
\begin{document}
\bstctlcite{IEEEexample:BSTcontrol}

%
\title{\emph{Cheetah}: Mixed Low-Precision Hardware \& Software Co-Design Framework for DNNs on the Edge}
%
%
%

\author{Hamed F. Langroudi,
        Zachariah Carmichael,
        David Pastuch,
        Dhireesha Kudithipudi
\thanks{Hamed. F. Langroudi, Zachariah Carmichael, David Pastuch, and Dhireesha Kudithipudi are with the Department
of Computer Engineering, Rochester Institute of Technology, Rochester,
NY, USA}}

%



\IEEEspecialpapernotice{(Preprint)}

\maketitle

\begin{abstract}
Low-precision DNNs \textcolor{black}{have been extensively explored in order} to reduce the size of DNN models for edge devices. Recently, the posit numerical format has shown promise for DNN data representation and compute with ultra-low precision $\in$ $[5..8]$ bits. However, previous studies were limited to studying posit for DNN inference only. In this paper, we propose the \emph{Cheetah} framework, which supports both DNN training and inference \textcolor{black}{using posits, as well as other commonly used formats}. Additionally, the framework is amenable for different quantization approaches and supports mixed-precision floating point and fixed-point numerical formats. \emph{Cheetah} is evaluated on \textcolor{black}{three datasets: MNIST, Fashion MNIST, and CIFAR-10}. 
\textcolor{black} {Results indicate that 16-bit posits outperform 16-bit floating point in DNN training. Furthermore, performing inference with [5..8]-bit posits improves the trade-off between performance and energy-delay-product over both [5..8]-bit float and fixed-point.}
\end{abstract}

\begin{IEEEkeywords}
Deep neural networks, low-precision arithmetic, posit numerical format 
\end{IEEEkeywords}

\IEEEpeerreviewmaketitle

\section{Introduction}

Edge computing is an emerging design paradigm that offers intelligence-at-the-edge \textcolor{black}{of} mobile networks, while addressing some of the shortcomings of cloud datacenters \cite{2018edge}. The nodes of the edges host the computing, storage, and communication capabilities, which \textcolor{black} {provide} on-demand learning for several applications\textcolor{black}{,} such as intelligent transportation, smart cities, \textcolor{black}{and} industrial robotics. Inherent characteristics of edge devices include low latency, reduced data movement cost, low communication bandwidth, and decentralized real-time processing \cite{shi2016edge,satyanarayanan2017emergence}. However, deploying intelligence-at-the-edge is a formidable challenge for several of the deep neural network (DNN) models. For instance, DNN inference with AlexNet requires $\sim${61 M} parameters and $\sim$1.4 \textcolor{black}{gigaFLOPS} \textcolor{black} {\cite{AlexNet}}. 
Moreover, the cost of the \textcolor{black} {multiply-and-accumulate (MAC) units, a fundamental DNN operation,} is non-trivial. \textcolor{black} {In a 45~nm CMOS process}, energy \textcolor{black} {consumption} doubles from 16-bit float\textcolor{black}{s} to 32-bit float\textcolor{black}{s} for addition and it increases by $\sim$4x for multiplication \textcolor{black} {\cite{horowitz20141}}. Memory access cost increases by $\sim${10x} from 8~\textcolor{black}{k} to 1~M memory size with 64-bit cache \cite{horowitz20141}. In general, there is a gap between memory storage, bandwidth, compute requirements\textcolor{black}{, and} energy consumption of today's DNN models and hardware resources available on edge devices \cite{xu2018scaling,facebookEdge}. 

\textcolor{black} {An apparent solution to address this gap is by compressing the size of the networks and \textcolor{black} {reduce} the computation requirements to match \textcolor{black} {putative} edge resources. Several groups have proposed compressed DNN models with new compute-and memory-efficient neural networks \cite{MobileNet,chen2019drop,MEC} and parameter-efficient neural networks, such as DNN pruning \cite{ren2018sbnet}, distillation \cite{zhou2018revisiting}, and low-precision arithmetic \cite{Jacob_2018_CVPR,IBM8}.} 

\textcolor{black}{Among these approaches to compress DNN models, low-precision arithmetic is noted for its ability to reduce memory capacity, bandwidth, latency, and energy consumption associated with MAC units in DNNs, and an increase in the level of data parallelism \cite{Jacob_2018_CVPR, hashemi2017, gysel2018}. For instance, DNN inference with compressed models, such as MobileNet with 8-bit fixed-point parameters, utilizes only \textcolor{black} {$\sim${4.2 M}} parameters and \textcolor{black} {$\sim${1.1} megaFLOPS} \cite{MobileNet}. While this alleviates \textcolor{black}{some} of the design constraints for the edge, DNN models must still run quickly with high accuracy for complex visual or video recognition tasks on-device. Therefore, a conflicting design constraint here is that the network's precision cannot compromise a DNN's overall performance. For instance, there is a ${\sim}10\%$ gap between the performance of low-precision DNN models (e.g, MobileNet with 8-bit fixed-point DNN parameters) and high-precision DNN models (e.g, MobileNet with 32-bit floating point DNN parameters) for real-time (30 FPS) classification on ImageNet data with a Snapdragon 835 LITTLE core \cite{Jacob_2018_CVPR}.}





\textcolor{black} {The ultimate goal of designing the low-precision DNN is reducing the hardware complexity of the high-precision DNN model such that it can be ported on to edge devices with performance similar to the high-precision DNN.} The hardware complexity and performance in low-precision DNNs rely heavily on the quantization approach and \textcolor{black}{the numerical format}. Prevailing techniques, such as complex vector quantization or hardware-friendly numerical formats, lead to undesirable hardware complexity or performance penalties \cite{guo2018survey, quantizingsurvey1}. 

To understand the correlation between hardware complexity and performance of low-precision neural networks for the edge, a hardware and software co-design framework is required. Previous studies have addressed this by proposing low-precision frameworks \cite{gysel2018,Jacob_2018_CVPR,IBM8,hashemi2017,Hamed2018,carmichael2019positron, carmichael2019performance,johnson2018rethinking}. However, the scope of these studies is limited, \textcolor{black}{as highlighted below:}

\begin{enumerate}
    \item None of the previous works explore the propriety of the posit numerical format for both DNN training and inference by comprehensive comparison with fixed and float formats  \cite{Hamed2018,carmichael2019positron,carmichael2019performance,johnson2018rethinking}.
    \item \textcolor{black}{There is a l}ack of comparison \textcolor{black}{between} the efficacy of quantization approaches\textcolor{black}{,} numerical formats\textcolor{black}{, and} the associated hardware complexity.
    \item In most of the previous works, the comparison across numerical formats are conducted for varying bit-widths (e.g. 32-bit floating point compared to 8-bit fixed-point \cite{hashemi2017}). Such comparisons do not offer insights on viability of \textcolor{black}{utilizing the} same bit-precision across numerical formats for a particular task.
\end{enumerate}

To address the gaps in previous studies, we are motivated to propose \emph{Cheetah} as a comprehensive hardware and software co-design framework to explore the advantage of low-precision for both DNN training and inference. The current version of \emph{Cheetah} supports three numerical formats (fixed-point, floating point, and posit), two quantization approaches (rounding and linear), and \textcolor{black}{two DNN models (feedforward neural networks and convolutional neural networks)}.

\section{Background}
\subsection{Deep Neural Network}
Deep neural networks (DNNs) \cite{lecun1998gradient} are artificial neural networks that are used for various tasks, such as classification, regression and prediction, by learning the correlation between examples from a corpus of data called training sets \cite{DeepLearningbook}. These networks are capable of learning a non-linear input-to-output mapping in either a supervised, unsupervised, or semi-supervised manner. \textcolor{black}{The DNN models contain a sequence of layers, each comprising a set of nodes. The connectivity between layers depends on the DNN architecture (e.g. globally connected in feedforward neural network or locally connected in convolutional neural network)}.

A major computation in a DNN node is the MAC operation. Specifically, a node in feedforward neural and convolutional neural network computes \eqref{eq:FeedForwardNetwork-Convolution} where $B$ indicates the bias vector, $W$ is the weight\textcolor{black}{s} tensor with numerical values that are associated with each connection, $A$ represents the activation vector as input values to each node, \textcolor{black} {$Y$ is the feature vector at the output of each node, and $N$ equals either the number of nodes for a feedforward neural network or the product of the $(C,R,S)$ filter parameters: the number of filter channels, the filter heights, and the filter weights, respectively, for a convolutional neural network.}
\textcolor{black} {
\begin{equation}\label{eq:FeedForwardNetwork-Convolution}
Y_j = B_j + \sum_{i=0}^{N} {A_{i} \times W_{ij}}    
\end{equation}
}


In a supervised learning scenario for all of these networks, \textcolor{black}{the} correctness of classification\textcolor{black}{s is given by} the distance between $Y$ and the desired output \textcolor{black}{as} calculated \textcolor{black}{by} $E_i$\textcolor{black}{, a} cost function with respect to the weights. Then, during training, the weights are learned through \textcolor{black} {stochastic gradient descent (SGD)} to minimize $E_i$ as given by \eqref{eq:GradientDescent}.

\begin{equation}\label{eq:GradientDescent}
\Delta W_{ij} = -\alpha \frac{\partial E_i}{\partial W_{ij}}  
\end{equation}
\subsection{Posit Numerical Format}

The posit, \textcolor{black} {a Type III unum}, is a new numerical format with tapered precision characteristic and was proposed as an alternative to \textcolor{black}{IEEE-754} floating format to represent real numbers \cite{gustafson2017beating}. Posit revamped the \textcolor{black}{IEEE-754} floating format and addressed complaints about Type I and Type II unums \cite{tichy2016unums}. Posits provides better accuracy, dynamic range, and program reproducibility than IEEE floating point. The essential advantage of posits is their capability to represent non-linearly distributed numbers in a specific dynamic range around 1 with maximum accuracy.
The value of a posit number is represented by \eqref{equ:equ00}, where $s$ represents the sign, $es$ and $fs$ represent the maximum number of bits allocated for the exponent and fraction, respectively, $e$ and $f$ indicate the exponent and fraction values, respectively, and $k$, as computed by (\ref{equ:equ01}), represents the regime value.
\begin{equation}
    x= 
\begin{cases}
    0, & \text{if } ({\tt 00...0})   \\
    NaR, & \text{if }  ({\tt 10...0}) \\
(-1)^{s}\times 2^{2^{es} \times k} \times 2^e \times \left(1+\frac{f}{2^{fs}} \right), & \text{otherwise}
\end{cases}
\label{equ:equ00}
\end{equation}
The regime bit-field is encoded based on the \emph{runlength} $m$ of identical bits $(r...r)$ terminated by either a \emph{regime terminating bit}~$\overline{r}$ or the end of the $n$-bit value. Note that there is no requirement to distinguish between negative and positive zero since only a single bit pattern ${\tt (00...0)}$ represents zero. Furthermore, instead of defining a \textit{NaN} for exceptional values and infinity by various bit patterns, a single bit pattern $({\tt 10...0})$, ``Not-a-Real'' ($NaR$), represents exception values and infinity. More details about the posit number format can be found in \cite{gustafson2017beating}.
\begin{equation}
    k= 
\begin{cases}
    -m , & \text{if } r= 0\\
     m-1, & \text{if } r= 1
\end{cases}
\label{equ:equ01}
\end{equation}

\section{Related Work}


\textcolor{black} {As lately as the 1980s}, low-precision arithmetic has been studied for shallow neural networks to reduce compute and memory complexity \textcolor{black}{for} training and inference without \textcolor{black}{sacrificing} performance\textcolor{black}{ \cite{graf1988vlsi,iwata1989artificial,hammerstrom1990vlsi, asanovic1991experimental}}. In some scenarios, \textcolor{black} {it} also improves the performance of training and inference \textcolor{black} {since the quantization noise generated from the use of low-precision parameters in shallow neural network acts as a regularization method \cite{asanovic1991experimental,bishop1995training}}.
The outcome of these studies indicate that \textcolor{black} {16- and 8-bit precision DNN parameters are} sufficient for training and inference on shallow networks \textcolor{black}{\cite{iwata1989artificial,hammerstrom1990vlsi,asanovic1991experimental}}. The capability of low-precision arithmetic is reevaluated in the deep learning era to reduce memory footprint and energy consumption during training and inference \cite{Courbariaux14,Gupta,micikevicius2017mixed,flexpoint2017,IBM8,mellempudi2019mixed,kalamkar2019study,gysel2018,hashemi2017,Microsoft2018,carmichael2019positron,carmichael2019performance,Hamed2018,johnson2018rethinking}. 
\subsection{\textcolor{black}{L}ow-\textcolor{black}{P}recision DNN \textcolor{black}{T}raining}
\textcolor{black} {Several of the previous studies have shown that to perform DNN training, either variants of low-precision block floating point (BFP), where a block of floating point DNN parameters used a shared exponent \cite{wilkinson1965rounding}, such as Flexpoint \cite{flexpoint2017} (16-bit fraction with 5-bit shared exponent for DNN parameters), or mixed-precision floating point (16-bit weights, activations, and gradients and 32-bit accumulators in the SGD weight update process) are sufficient to maintain similar performance as 32-bit high-precision floating point. For instance, Courbariaux \textit{et al.} trained a low-precision DNN on the MNIST, CIFAR-10, and SVHN datasets with the floating point, fixed-point, and BFP numerical formats \cite{Courbariaux14}. They demonstrate that BFP is the most suitable choice for low-precision training due to variability between the dynamic range and precision of DNN parameters \cite{Courbariaux14}. Following this work, Koster \textit{et al.} proposed the Flexpoint numerical format and a new algorithm called Autoflex to automatically predict the optimal shared exponents for DNN parameters in each iteration of SGD by statistically analyzing the values of DNN parameters in previous iterations \cite{flexpoint2017}.} 

\textcolor{black}{Aside from managing the shared exponent in the BFP numerical format, Narang \textit{et al.} used mixed-precision floating point \cite{micikevicius2017mixed}. They used a 16-bit floating point to represent weights, activations, and gradients to perform forward and backward passes. To prevent accuracy loss caused by underflow in the product of learning rate and gradients with \eqref{eq:GradientDescent} in 16-bit floating point, the weights are updated in 32-bit floating point. Additionally, to prevent gradients with very small magnitude from becoming zero when represented by 16-bit float, a new loss scaling approach is proposed \cite{micikevicius2017mixed}.}

\textcolor{black}{Recently, Wang \textit{et al.} and Mellempudi \textit{et al.} reduce the bit-precision required to represent weights, activations, and gradients to 8-bit by exhaustively analyzing DNN training parameters \cite{IBM8,mellempudi2019mixed}. Even in \cite{mellempudi2019mixed}, a new chunk-based addition is presented to solve the truncation issue caused by addition of large- and small-magnitude numbers and thus the number of bits demanded for accumulator and weight updates is reduced to 16-bits. To prevent the requirement of the loss scaling in mixed-precision floating point, Kalamkar \textit{et al.} \cite{kalamkar2019study} proposed the brain floating point (BFLOAT-16) half-precision format with similar dynamic range (7-bit exponent) and less precision (8-bit fraction) compared to 32-bit floating point. The same dynamic range between BFLOAT-16 and 32-bit floating point reduces the conversion complexity between these two formats in DNN training. In training a ResNet model on the ImageNet dataset, BFLOAT-16s achieve the same performance as 32-bit floating point.}

\subsection{\textcolor{black}{L}ow-\textcolor{black}{P}recision DNN \textcolor{black}{I}nference}

\textcolor{black} {The performance of DNN inference without retraining is more robust to the noise that is generated from low-precision DNN parameters as the DNN parameters during inference are static; several groups have demonstrated that either 8-bit BFP or 8-bit fixed-point, coupled with linear quantization, are adequate to represent weights and activations without significantly degrading performance yielded with 32-bit floating point. Note that the accumulation bit-width is selected to be 32~bits to preserve accuracy in performing, in general, thousands of additions in the MAC operations. For instance, Gysel \textit{et al.} demonstrate that an 8-bit block floating point for representing weights and activations, 8-bit multipliers, and 32-bit accumulation results in $<$1\% accuracy loss on AlexNet with the ImageNet corpus \cite{gysel2018}. Following this work, Hashemi \textit{et al.} introduce low-precision DNN inference networks to better understand the impact of numerical formats on the energy consumption and performance of DNNs \cite{hashemi2017,gysel2018}. For instance, performing inference on AlexNet with the 8-bit fixed-point format yields a $6\times$ improvement in energy consumption over 32-bit fixed-point for the CIFAR-10 dataset \cite{hashemi2017}. Chung \textit{et al.} proposed the Brainwave accelerator using 8-bit block floating point with a 5-bit exponent to classify ImageNet dataset on ResNet-50 with $<$2\% accuracy loss \cite{Microsoft2018}. However, the scaling factor parameter in the block floating point numerical format needs to be updated according to the DNN parameter statistics, thus increasing the computational complexity of inference.}

\textcolor{black} {To alleviate this problem, researchers have used posits in DNNs \cite{carmichael2019positron,carmichael2019performance,Hamed2018,johnson2018rethinking}. Posits represent numbers more accurately around $\pm$1 and less accurately for very small and large numbers, unlike the uniform precision of the floating point numerical format \cite{dedinechin}.
This characteristic of posits arises from its tapered precision and suits the distribution of DNN parameters well \cite{gustafson2017beating, Hamed2018}. For instance, Langroudi \textit{et al.} explored the efficacy of posits for representing DNN weights and have shown that it is possible to achieve a loss in accuracy within $<$1\% on the AlexNet and ImageNet corpora with weight representation at 7-bit \cite{Hamed2018}.
They also demonstrate that posits have a 30\% less voracious memory footprint than fixed-point for multiple DNNs while maintaining a $<$1\% drop in accuracy. However, in the work, the 7-bit posit quantized weights are converted to 32-bit floats, limiting the posit numerical format for memory storage only.}

\textcolor{black}{To take full advantage of the posit numerical format, Carmichael \textit{et al.} proposed the Deep Positron DNN accelerator which employs the posit numerical format to represent weights and activations combined with an FPGA soft core for $\leq$8-bit precision exact-MAC operations \cite{carmichael2019positron,carmichael2019performance}. They demonstrate that 8-bit posits outperform 8-bit fixed-point and floating point on low-dimensional datasets, such as Iris \cite{fisher1936use}. Following these works, most recently, Jeff Johnson proposed a log float format as a combination of the posit numerical format and exact log-linear multiply-add (ELMA), which is the logarithmic version of the exact MAC operation. This work shows that it is possible to classify ImageNet with the ResNet DNN architecture with $<$1\% accuracy degradation \cite{johnson2018rethinking}}.

\begin{figure*}
\centering
\includegraphics[width=.75\linewidth]{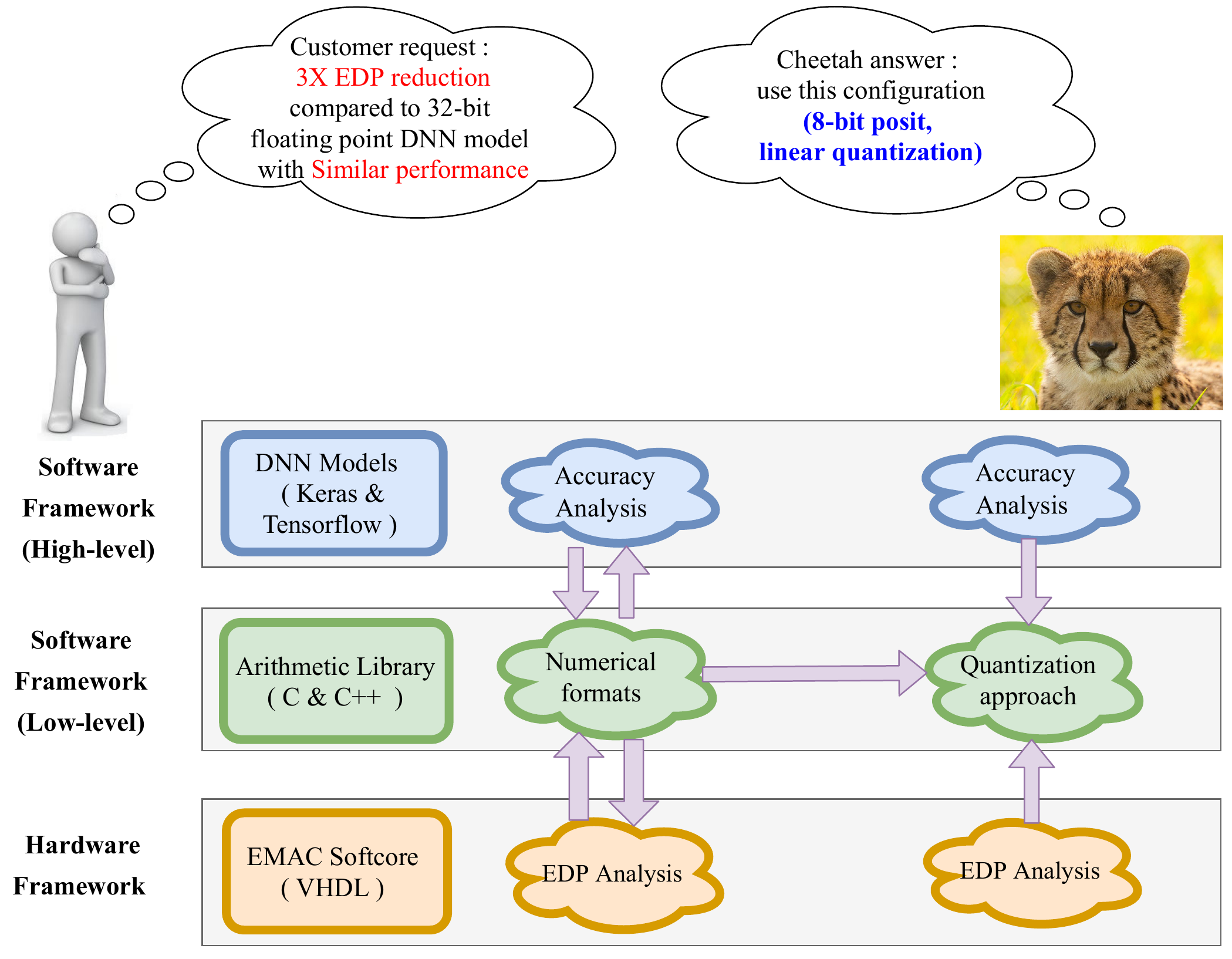}
\caption{The \emph{Cheetah} High-level Hardware \& Software Co-design framework for DNNs on the edge. EDP: Energy-Delay Product.}
\label{fig:HighLevelCheetah}
\vspace{-5mm}
\end{figure*}

\textcolor{black} {This research builds on these earlier studies \cite{carmichael2019positron,carmichael2019performance,Hamed2018,johnson2018rethinking} and extends low-precision arithmetic to both DNN training and DNN inference with different quantization approaches for both feedforward and convolution neural networks on various datasets.}


\section{Proposed Framework}
\begin{figure*}
\centering
\includegraphics[width=.5\linewidth]{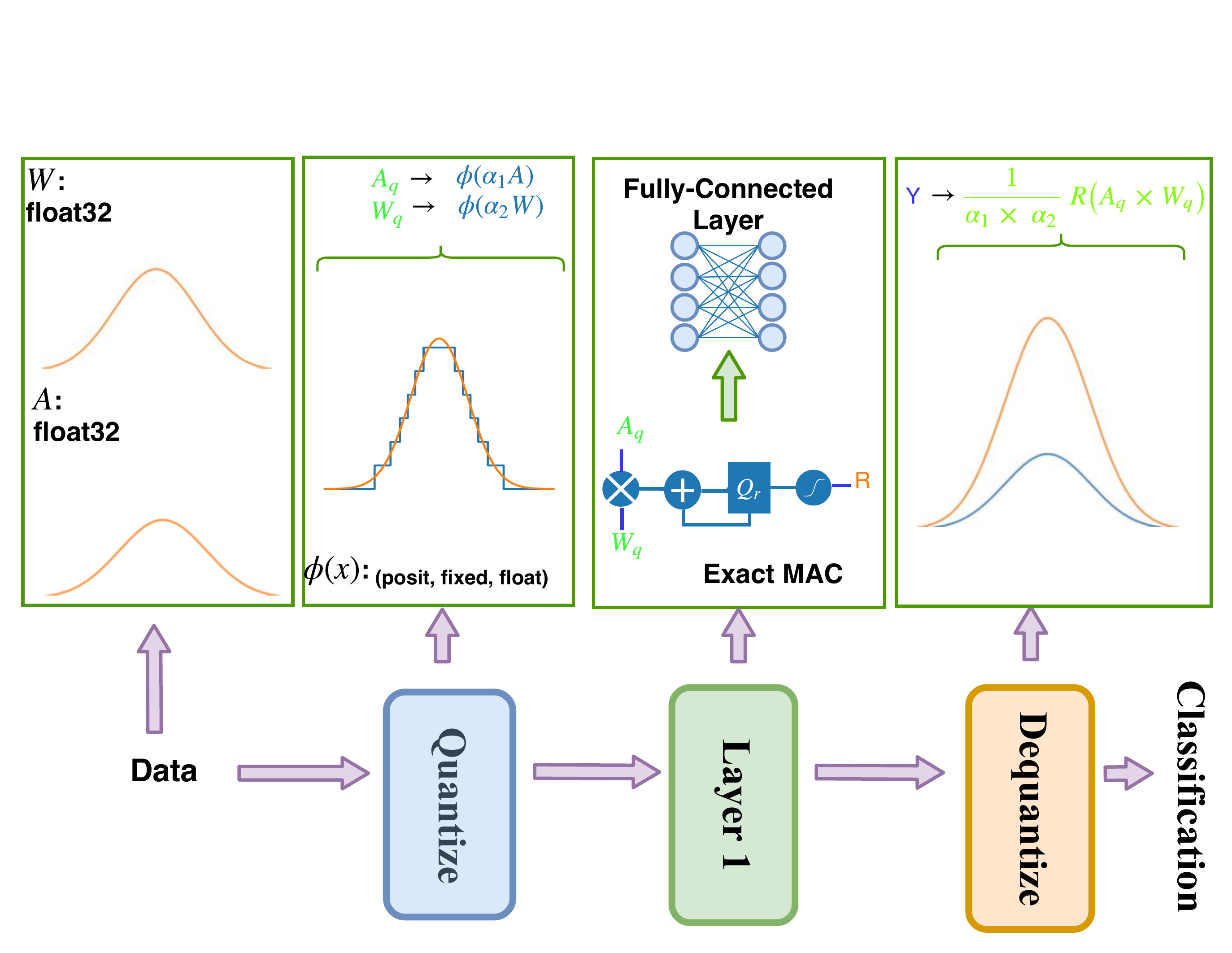}
\caption{The \emph{Cheetah} \textcolor{black} {software} framework for feedforward neural networks with one hidden layer. The framework scales to any DNN architecture.}
\label{fig:Software-Cheetah}
\vspace{-5mm}
\end{figure*}

\textcolor{black} {The \emph{Cheetah} framework, shown in Fig. \ref{fig:HighLevelCheetah}, comprises a two-level software component and a single-level hardware component.}
The software framework is used to evaluate the performance of various numerical formats and quantization approaches by emulating low-precision DNN training and inference. The hardware framework is a soft-core implemented on FPGA and used for evaluating hardware characteristics of the MAC (multiply-and-accumulate) operations as a fundamental computation in DNN models coupled with various quantization techniques. \textcolor{black}{For each level, two optimization stages are considered to convert the baseline DNN model with 32-bit high-precision floating point with soft-core MACs to a low-precision DNN model with either posit, floating point, or fixed-point arithmetic soft-core exact-MACs (EMACs). This optimization is performed iteratively, reducing the bit-precision by one at each step; the performance degradation and hardware complexity reduction achieved by a numerical format in both DNN training and inference is computed and compared with the specified design constraints (e.g. 3$\times$ EDP reduction with similar performance). This iterative process is repeated for the next numerical format after one of the design constraints is violated. Essentially, \emph{Cheetah} approximates the optimal bit-width for each numerical format based on the performance and hardware complexity constraints. Note that there is a priority between optimization approaches; the numerical format parameter has a higher precedence in the optimization process. This design decision is made to limit the search space and the hardware complexity overhead of the quantization approaches. In performing DNN inference,} the current version of \emph{Cheetah} supports three low-precision numerical formats (fixed-point, floating point and posit), two quantization approaches (rounding and linear), and two DNN model\textcolor{black}{s} (feedforward and convolutional neural networks). \textcolor{black}{To perform DNN training on feedforward neural networks, \emph{Cheetah} supports two numerical formats (floating point and posit) with 32-bit and 16-bit precision. For brevity, the
architecture 
explained here is based on single hidden layer feedforward neural network training and inference with the posit numerical format for both rounding and linear quantization approaches, as shown in Fig. \ref{fig:Software-Cheetah}.}

\subsection{Software Design and Exploration}
\textcolor{black}{In emulating feedforward and convolutional DNNs, the output of each layer $Y$ is calculated as in \eqref{eq:FeedForwardNetwork-Convolution-Quantized}}
\textcolor{black}{
\begin{equation}
Y_j{=}B_j{+}\frac{1}{\alpha_1{\times}\alpha_2}{\times}
\left(
    \sum_{i}^{N}
    \left[ Q(\alpha_1{\times}A_i) \right]
    {\times}
    \left[ Q(\alpha_2{\times}W_{ij}) \right]
\right)
\label{eq:FeedForwardNetwork-Convolution-Quantized}
\end{equation}
}

\noindent
where $\alpha_1$ and $\alpha_2$ are scale factors, $B_i$ is the bias term, $A_i$ is the activation vector, $W_{ij}$ is the weight matrix, $N$ indicates the number of \textcolor{black}{MAC operations, and $Q(\cdot)$ is the quantization function}. \textcolor{black}{First, the feedforward or convolutional neural network} is trained by either 32- \textcolor{black}{or} 16-bit floating point or \textcolor{black}{p}osit \textcolor{black}{numbers} as shown by Fig. \ref{eq:FeedForwardNetwork-Convolution-Quantized}. To perform DNN inference, the 32-bit floating point high-precision learned weights \textcolor{black}{and} 32-bit floating point high-precision activations are quantized to either $n$-bit low-precision fixed-point\textcolor{black}{,} floating point\textcolor{black}{,} or posit \textcolor{black}{numbers} ($n \leq 8$).


In the quantization procedure, the values of $\alpha_1$ and $\alpha_2$ are dependent on the quantization approach. To perform rounding quantization, $\alpha_1$ and $\alpha_2$ are both set to 1 and the 32-bit high-precision floating point values that lie outside dynamic range of one of the low-precision posit numerical formats (e.g. 8-bit posit) are clipped appropriately to either the format's maximum or minimum. During quantization by rounding, a value that is interleaved between two arbitrary numbers \textcolor{black}{is} rounded to the nearest \textcolor{black}{number}. To perform linear quantization, the activations and weights are quantized to the range $[-\beta,\beta]$ by calculating $\alpha_1= \frac{\beta}{\textup{Max}(A_i)}$ and setting $\alpha_2= \frac{2 \beta}{\textup{Max}(W_i) - \textup{Min}(W_i)}$. 

In the next step, \textcolor{black} {the MAC operation is employed} to calculate $Y_i$. To minimize arithmetic error,
the MAC operation in this paper is calculated using the EMAC algorithm \cite{carmichael2019positron}. In the EMAC, to preserve precision in computing the products, the posit weights and activations are multiplied in a posit format without truncation or rounding at the end of multiplications. To avoid rounding during accumulation, the products are stored in a wide register, or \textit{quire} in the posit literature, with a width given by \eqref{eq:kulisch_width}. The products are then converted to the fixed-point format $FX_{(m_k,n_k)}$, where
$m_k = 2 ^ {es + 1} \times (n - 2) + 2 + \lceil{\log_2 (N_{op})}\rceil$ \textcolor{black} {is the exponent bit-width} and $n_k = 2 ^ {es + 1} \times (n - 2)$ \textcolor{black}{is the fraction bit-width}.
Finally, the $N_{op}$ fixed-point products are accumulated and the result is descaled in linear quantization, again using $\alpha_1$ and $\alpha_2$, and converted back to posit.


\begin{equation}\label{eq:kulisch_width}
	w_q = \lceil{\log_2 (N_{op})} \rceil + 2^{\,es+2} \times (n-2) + 2 
\end{equation}

\hspace{0pt}  
\begin{breakablealgorithm}
  \caption{Posit DOT operation for $n$-bit inputs each with $es$ exponent bits \cite{carmichael2019positron}}\label{alg:posit_edp}
  \begin{algorithmic}[1]
  	\begingroup
      \small
      \setlength{\thinmuskip}{2mu}
      \setlength{\medmuskip}{3mu plus 1.5mu minus 3mu}
      \setlength{\thickmuskip}{3.5mu plus 3.5mu}
      \Procedure{PositDOT}{$\tt{weight,activation}$}
        \State ${\tt{sign_w}}, {\tt{reg_w}}, {\tt{exp_w}}, {\tt{frac_w}} \gets \text{\headersty{Decode}}({\tt{weight}})$
        \State ${\tt{sign_a}}, {\tt{reg_a}}, {\tt{exp_a}}, {\tt{frac_a}} \gets \text{\headersty{Decode}}({\tt{activation}})$
        \State ${\tt{sf_w}} \gets \{{\tt{reg_w}}, {\tt{exp_w}}\}$\Comment{Gather scale factors}
        \State ${\tt{sf_a}} \gets \{{\tt{reg_a}}, {\tt{exp_a}}\}$
        \algrule \hspace{-1.75mm}\textbf{Multiplication}
        \State ${\tt{sign_{mult}}} \gets {\tt{sign_w}} \oplus {\tt{sign_a}}$
        \State ${\tt{frac_{mult}}} \gets {\tt{frac_w}} \times {\tt{frac_a}}$
        \State ${\tt{ovf_{mult}}} \gets {\tt{frac_{mult}}}[{\tt{MSB}}]$\Comment{Adjust for overflow}
        \State ${\tt{normfrac_{mult}}} \gets {\tt{frac_{mult}}} \gg {\tt{ovf_{mult}}}$
        \State ${\tt{sf_{mult}}} \gets {\tt{sf_{w}}} + {\tt{sf_{a}}} + {\tt{ovf_{mult}}}$
        \algrule \hspace{-1.75mm}\textbf{Accumulation}
        \State ${\tt{fracs_{mult}}} \gets {\tt{sign_{mult}}}~?~{\unaryminus\tt{frac_{mult}}}:{\tt{frac_{mult}}}$
        \State ${\tt{sf_{biased}}} \gets {\tt{sf_{mult}}} + bias$\Comment{Bias the scale factor}
        \State ${\tt{fracs_{fixed}}} \gets {\tt{fracs_{mult}}} \ll {\tt{sf_{biased}}}$\Comment{Shift to fixed}
        \State ${\tt{sum_{quire}}} \gets {\tt{fracs_{fixed}}} + {\tt{sum_{quire}}}$\Comment{Accumulate}
        \algrule \hspace{-1.75mm}\textbf{Fraction \& SF Extraction}
        \State ${\tt{sign_{quire}}} \gets {\tt{sum_{quire}}}[{\tt{MSB}}]$
        \State ${\tt{mag_{quire}}} \gets {\tt{sign_{quire}}}~?~{\unaryminus\tt{sum_{quire}}}:{\tt{sum_{quire}}}$
        \State ${\tt{zc}} \gets \text{\headersty{LeadingZerosDetector}}({\tt{mag_{quire}}})$
        \State ${\tt{frac_{quire}}} \gets {\tt{mag_{quire}}}[2{\times}(n\unaryminus 2\unaryminus es)\unaryminus 1{+}{\tt{zc}}:{\tt{zc}}]$
        \State ${\tt{sf_{quire}}} \gets {\tt{zc}} \unaryminus bias$
        \algrule \hspace{-1.75mm}\textbf{Convergent Rounding \& Encoding}
        \State ${\tt{nzero}} \gets |{\tt{frac_{quire}}}$
        \State ${\tt{sign_{sf}}} \gets {\tt{sf_{quire}}}[{\tt{MSB}}]$
    	\State ${\tt{exp}} \gets {\tt{sf_{quire}}}[es\unaryminus 1:0]$\Comment{Unpack scale factor}
        \State ${\tt{reg_{tmp}}} \gets {\tt{sf_{quire}}}[{\tt{MSB}}\unaryminus1:es]$
        \State ${\tt{reg}} \gets {\tt{sign_{sf}}}~?~\unaryminus{\tt{reg_{tmp}}}:{\tt{reg_{tmp}}}$
        \State ${\tt{ovf_{reg}}} \gets {\tt{reg}}[{\tt{MSB}}]$\Comment{Check for overflow}
        \State ${\tt{reg_f}} \gets {\tt{ovf_{reg}}}~?~\{\{\lceil\log_2(n)\rceil\unaryminus 2 \{{\tt{1}}\}\}), {\tt{0}}\} : {\tt{reg}}$
        \State ${\tt{exp_f}} \gets ({\tt{ovf_{reg}}}|{\sim}{\tt{nzero}}|{(\tt{\&{\tt{reg_f}}}}))~?~\{es\{{\tt{0}}\}\}:{\tt{exp}}$
        \State ${\tt{tmp1}} \gets \{{\tt{nzero}}, {\tt{0}}, {\tt{exp_f}}, {\tt{frac_{quire}}}[{\tt{MSB}}\unaryminus 1:0],$ \par
        $~~~~~~~~~~\{n\unaryminus 1 \{{\tt{0}}\}\}\}$
        \State ${\tt{tmp2}} \gets \{{\tt{0}}, {\tt{nzero}}, {\tt{exp_f}}, {\tt{frac_{quire}}}[{\tt{MSB}}\unaryminus 1:0],$ \par
        $~~~~~~~~~~\{n\unaryminus 1 \{{\tt{0}}\}\}\}$
        \State ${\tt{ovf_{regf}}} \gets \&{\tt{reg_f}}$
        \If {${\tt{ovf_{regf}}}$}
        	\State ${\tt{shift_{neg}}} \gets {\tt{reg_{f}}} - 2$
			\State ${\tt{shift_{pos}}} \gets {\tt{reg_{f}}} - 1$
		\Else
			\State ${\tt{shift_{neg}}} \gets {\tt{reg_{f}}} - 1$
			\State ${\tt{shift_{pos}}} \gets {\tt{reg_{f}}}$
		\EndIf
        \State ${\tt{tmp}} \gets {\tt{sign_{sf}}}~?~{\tt{tmp2}} \gg {\tt{shift_{neg}}} : {\tt{tmp1}} \gg {\tt{shift_{pos}}}$
        \State ${\tt{lsb}}, {\tt{guard}} \gets {\tt{tmp}}[{\tt{MSB}}\unaryminus (n\unaryminus 2):{\tt{MSB}}\unaryminus (n\unaryminus 1)]$
        \State ${\tt{round}} \gets {\sim}({\tt{ovf_{reg}}}|{\tt{ovf_{regf}}})~? $ \par
        $~~~~~~~~~~~~~~~~~~(~{\tt{guard}}~\&~({\tt{lsb}}~|~(|{\tt{tmp}}[{\tt{MSB}}\unaryminus n : 0]))~) : {\tt{0}}$
        \State ${\tt{result_{tmp}}} \gets {\tt{tmp}}[{\tt{MSB}} : {\tt{MSB}}\unaryminus n{+}1] {+} {\tt{round}}$
		\State ${\tt{result}} \gets {\tt{sign_{quire}}}~?~\unaryminus{\tt{result_{tmp}}} : {\tt{result_{tmp}}}$
        \State \textbf{return} ${\tt{result}}$
      \EndProcedure
    \endgroup
  \end{algorithmic}
\end{breakablealgorithm}

\begin{figure*}
  \centering
  \includegraphics[width=\linewidth]{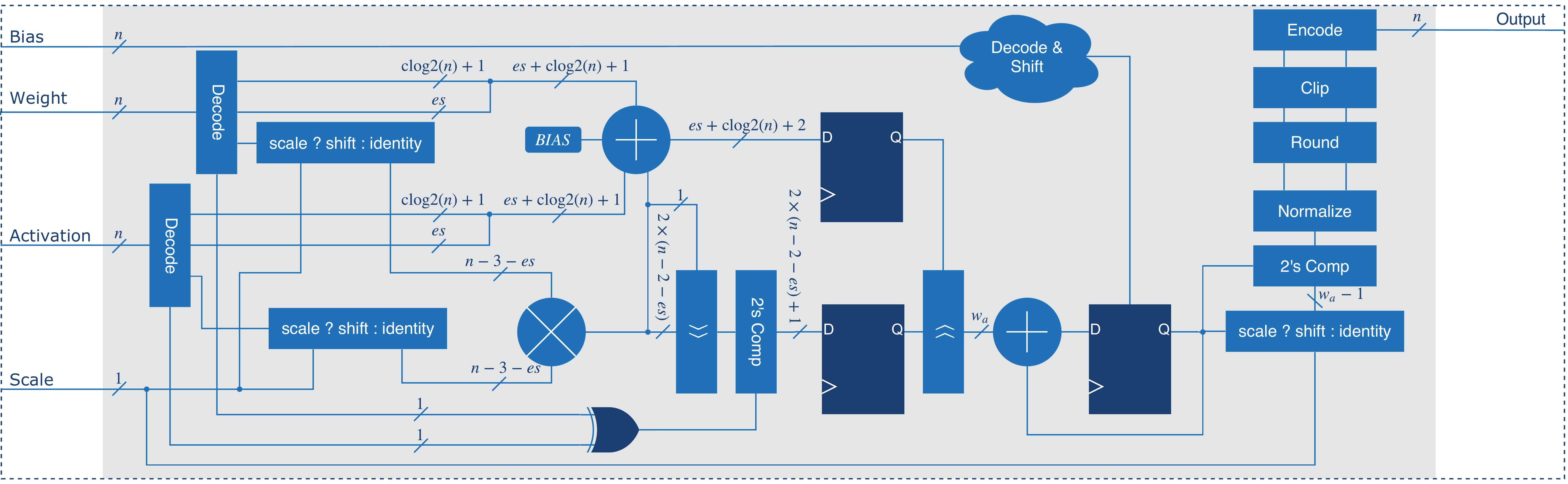}
  \caption{A parameterized ($n$ total bits, $es$ exponent bits) FPGA soft core design of the posit exact multiply-and-accumulate (EMAC) operation \cite{carmichael2019positron}.}
  \label{fig:posit_mac}
\end{figure*}

\subsection{Hardware Framework}
The MAC operation, as introduced as the fundamental DNN operation, calculates the weighted sum of a set of inputs. In many implementations, this operation is inexact, i.e. arithmetic error grows due to iterative rounding and truncation. The EMAC mitigates this concern by adapting the concept of the Kulisch accumulator \cite{kulisch2013computer}\textcolor{black} {. The error due to rounding} is deferred until after the accumulation of all products, which low-precision arithmetic further benefits from. In the EMAC, as mentioned \textcolor{black}{beforehand,} the fixed-point values of $N_{op}$ products are accumulated in a wide register sized as \textcolor{black}{given by \eqref{eq:kulisch_width}}. The posit EMAC, illustrated by Fig. \ref{fig:posit_mac}, is parameterized by $n$, the bit-width, and $es$, the number of exponential bits. ``NaR'' is not considered as posits do not overflow or underflow and all DNN parameters and data are real numbers. Algorithm \ref{alg:posit_edp} describes the bitwise operation of the EMAC dot product. \textcolor{black} {Each EMAC is pipelined into three stages: multiplication, accumulation, and rounding.} \textcolor{black}{For further details on EMACs and the exact dot product, we suggest reviewing \cite{kulisch2013computer, carmichael2019positron, carmichael2019performance}}.

\section{Simulation Results \& Analysis}
The \emph{Cheetah} software is implemented in the Keras \cite{keras} and TensorFlow \cite{tensorflow2015} frameworks. Rounding quantization, linear quantization\textcolor{black}{,} and the EMAC operations \textcolor{black}{with} [5,32]-bit precision \textcolor{black}{f}ixed-point, floating point\textcolor{black}{,} and posit \textcolor{black}{numbers} for DNN inference and \textcolor{black}{\{16, 32\}}-bit \textcolor{black}{f}loating point and posit \textcolor{black}{numbers} for DNN training are extended to these frameworks via software emulation. To reduce the search space of \textcolor{black}{t}he $\alpha_1$ and $\alpha_2$ parameters, \textcolor{black} {$\beta$ is selected from $\{1,2,4,8\}$} which still provides\textcolor{black}{, on average,} a wide coverage \textcolor{black} {($\sim$82\%)} of the dynamic range of \textcolor{black}{each} numerical format\textcolor{black}{, as shown in Table \ref{tab:Whybeta}}.

\begin{table}[H]
\caption{\textcolor{black} {The dynamic range coverage of $\leq$8-bit posit, floating point, and fixed-point numerical formats. The percentages are calculated without considering (NaR), infinity, and ``Not-a-Number'' (NaN) values.}}\label{tab:Whybeta}
\centering
\ra{1.2}
\begin{threeparttable}
\begin{tabular}{@{}cc@{}}
    \toprule
    Format   & Dynamic Range $\leq$8-bit \\
    \midrule
    Posit ($es$=0)& 94.12\% \\
    Posit ($es$=1)& 81.57\% \\
    Posit ($es$=2)& 69.02\% \\
    Float ($w_e$=4) & 66.66\%  \\
    Float ($w_e$=3) & 85.71\% \\
    Fixed-point ($n_k$=4) & 100.0\% \\
    \bottomrule
\end{tabular}
\end{threeparttable}
\end{table}

\begin{table*}[ht!]
\caption{Specifications of the benchmark tasks and performance on a baseline 32-bit floating point network}
\label{table:Spec-32bit-accuracy}
\centering
\ra{1.2}
\begin{threeparttable}
\begin{tabular}{@{}cccccc@{}} 
    \toprule
     Dataset 
    & Layers\tnote{1}  & \# Parameters & 
    \# EMAC Ops\tnote{2}& Memory & 
    Accuracy \\
    \midrule
    \multirow{2}{*}{MNIST} 
    & 4 FC & 0.34 M & 0.78 k & 1.34 MB & 98.46\%  \\
    & 2 Conv, 2 FC, 1 PL & 1.40 M & 58.7 k & 5.84 MB & 99.32\%  \\
    \multirow{2}{*}{Fashion-MNIST} 
    & 4 FC & 0.34 M & 0.78 k & 1.34 MB
    & 89.51\%  \\
    &  2 Conv, 3 FC, 2 PL, 1 BN & 1.88 M & 69.8 k & 7.77 MB & 92.54\%  \\
    \multirow{1}{*}{CIFAR-10}
    &  7 Conv, 1 FC, 3 PL  & 0.95 M & 312.6 k & 6.23 MB & 81.37\%  \\
    \bottomrule
    
\end{tabular}
\begin{tablenotes}
\item[1]Conv: 2D convolutional layer; FC: fully-connected layer; PL: max/avg. pooling layer; BN: batch normalization layer.
\item[2]The number of EMAC operations for a single sample.
\end{tablenotes}
\end{threeparttable}
\vspace{-0.5mm}
\end{table*}

\begin{table*}[ht!]
\vspace*{0.05 cm}
\caption{\emph{Cheetah} accuracy on three datasets with [5..8]-bit precision compared to fixed and float (\textcolor{black}{r}espective best results are when posit has $es \in \{0, 1, 2\}$ and floating point with exponent bit-width $w_e \in \{3, 4\}$).}\label{table:Accuracy-Result}
\centering
\ra{1.3}
\begin{threeparttable}
\resizebox{\linewidth}{!}{
\begin{tabular}{@{}ccccccccccccccccc@{}} 
 \toprule
    \multirow{2}{*}{Dataset} & \multirow{2}{*}{DNN} && \multicolumn{4}{c}{Posit} && \multicolumn{4}{c}{Float} && \multicolumn{4}{c}{Fixed} \\
    \cmidrule{4-7}\cmidrule{9-12}\cmidrule{14-17}
     &&& 8-bit & 7-bit & 6-bit & 5-bit && 8-bit & 7-bit & 6-bit & 5-bit && 8-bit & 7-bit & 6-bit & 5-bit \\
    \midrule
    \multirow{2}{*}{MNIST} & FC && \textbf{98.45}\% & \textbf{98.39}\% & \textbf{98.37}\% & \textbf{98.30}\% && 98.42\% & 98.39\% & 98.33\% & 93.91\% && 98.31\% & 97.95\% & 97.87\% & 97.88\% \\[-0.5ex]
    & Conv && \textbf{99.35}\% & \textbf{99.33}\% & \textbf{99.20}\% & \textbf{98.94}\% && 99.34\% & 99.25\% & 99.12\% & 92.27\% && 99.18\% & 97.14\% & 97.08\% & 96.96\% \\
    Fashion & FC && \textbf{89.59}\% & \textbf{89.44}\% & \textbf{89.24}\% & \textbf{88.14}\% && 89.56\% & 89.36\% & 88.92\% & 83.00\% && 89.16\% & 87.27\% & 85.20\% & 83.97\% \\[-0.5ex]
    MNIST & Conv && \textbf{92.70}\% & \textbf{92.60}\% & \textbf{91.64}\% & \textbf{88.92}\% && 92.63\% & 92.22\% & 89.58\% & 68.21\% && 89.59\% & 88.63\% & 85.31\% & 83.46\% \\
    \multirow{1}{*}{CIFAR-10} & Conv && \textbf{80.40}\% & \textbf{76.90}\% &  \textbf{68.51}\% & \textbf{41.33}\% && 79.75\% & 76.09\% & 53.68\% & 12.83\% && 24.27\% & 17.43\% & 12.54\% & 9.71\%\\
    \bottomrule
\end{tabular}
}
\end{threeparttable}
\end{table*}

\begin{figure*}[!ht]
\centering
\begin{subfigure}{.075\linewidth}
  \centering
  \caption{}
\end{subfigure}%
\begin{subfigure}{.30\linewidth}
  \centering
  \includegraphics[width=\linewidth]{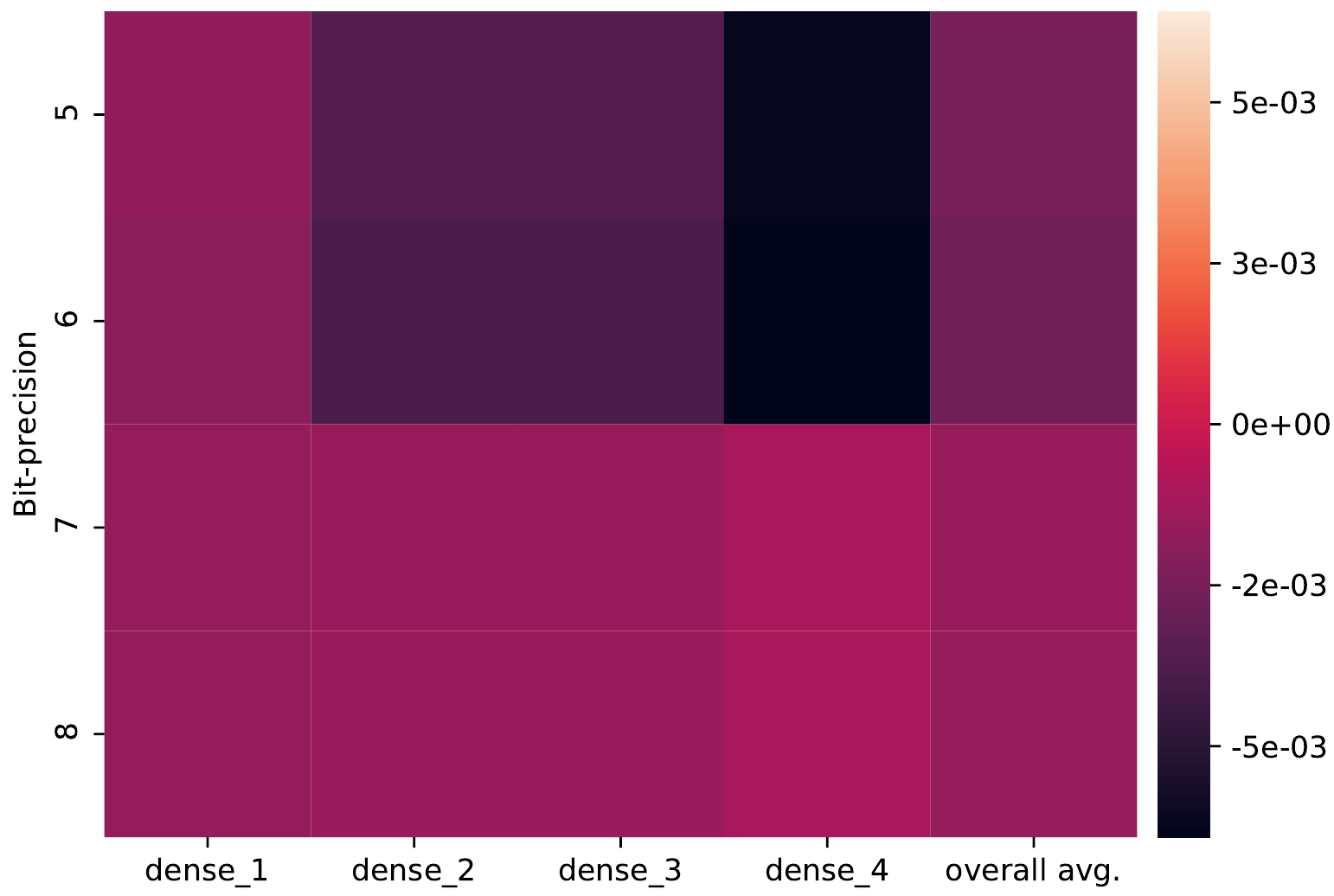}
\end{subfigure}%
\begin{subfigure}{.075\linewidth}
  \centering
  \caption{}
\end{subfigure}%
\begin{subfigure}{.30\linewidth}
  \centering
  \includegraphics[width=\linewidth]{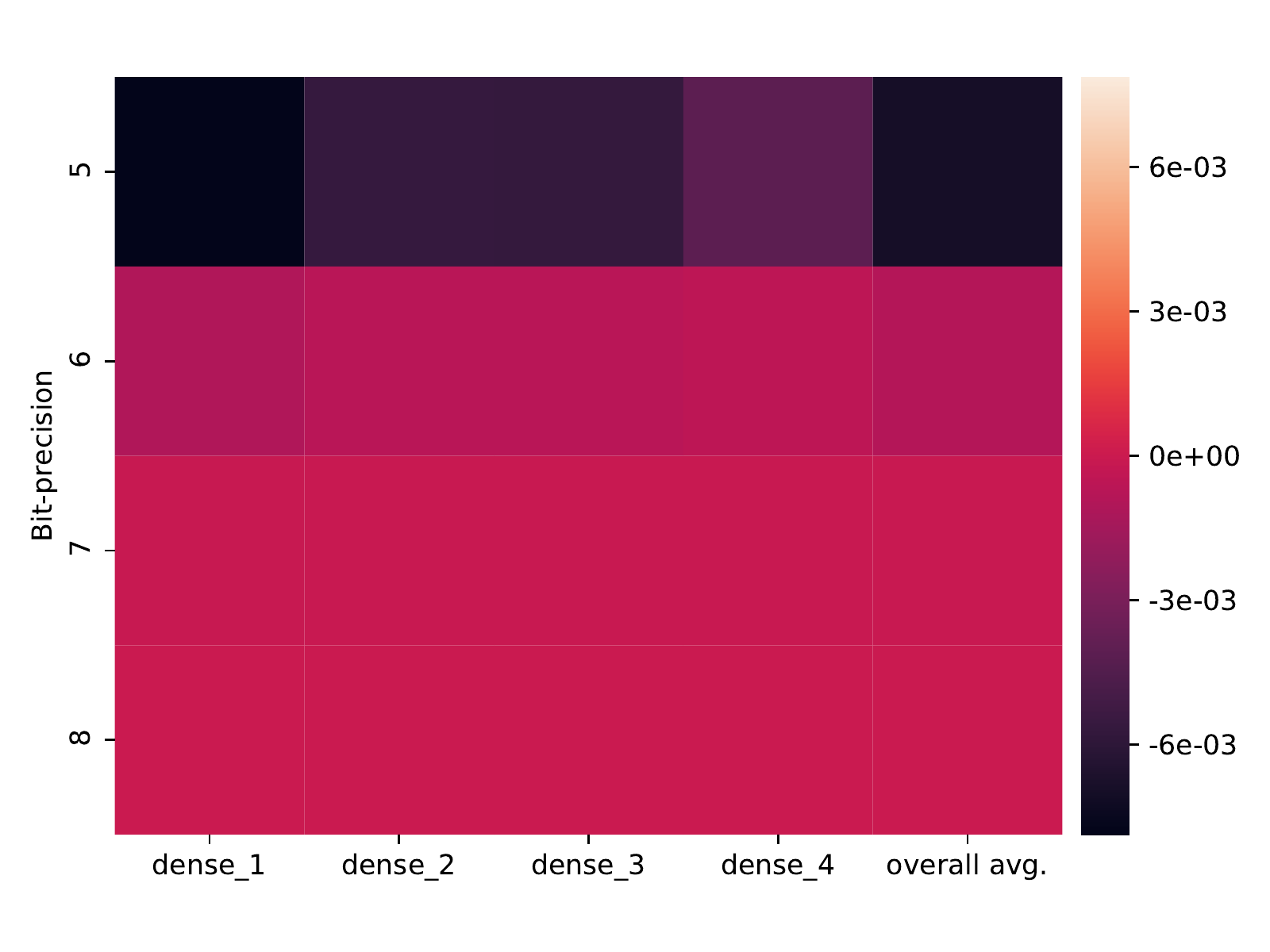}
\end{subfigure}\vskip\baselineskip\vspace{-3mm}
\begin{subfigure}{.075\linewidth}
  \centering
  \caption{}
\end{subfigure}%
\begin{subfigure}{.30\linewidth}
  \centering
  \includegraphics[width=\linewidth]{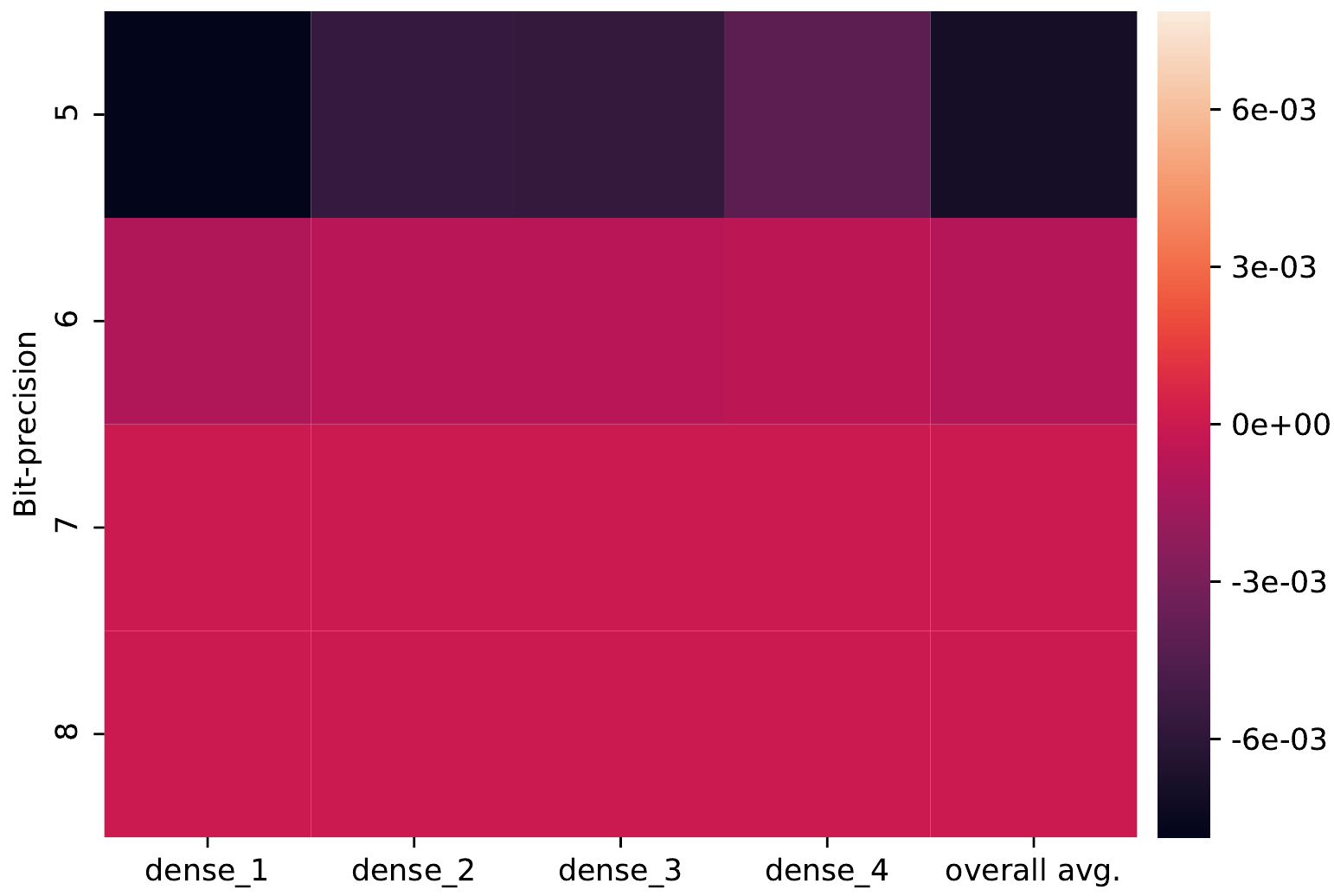}
\end{subfigure}%
\begin{subfigure}{.075\linewidth}
  \centering
  \caption{}
\end{subfigure}%
\begin{subfigure}{.30\linewidth}
  \centering
  \includegraphics[width=\linewidth]{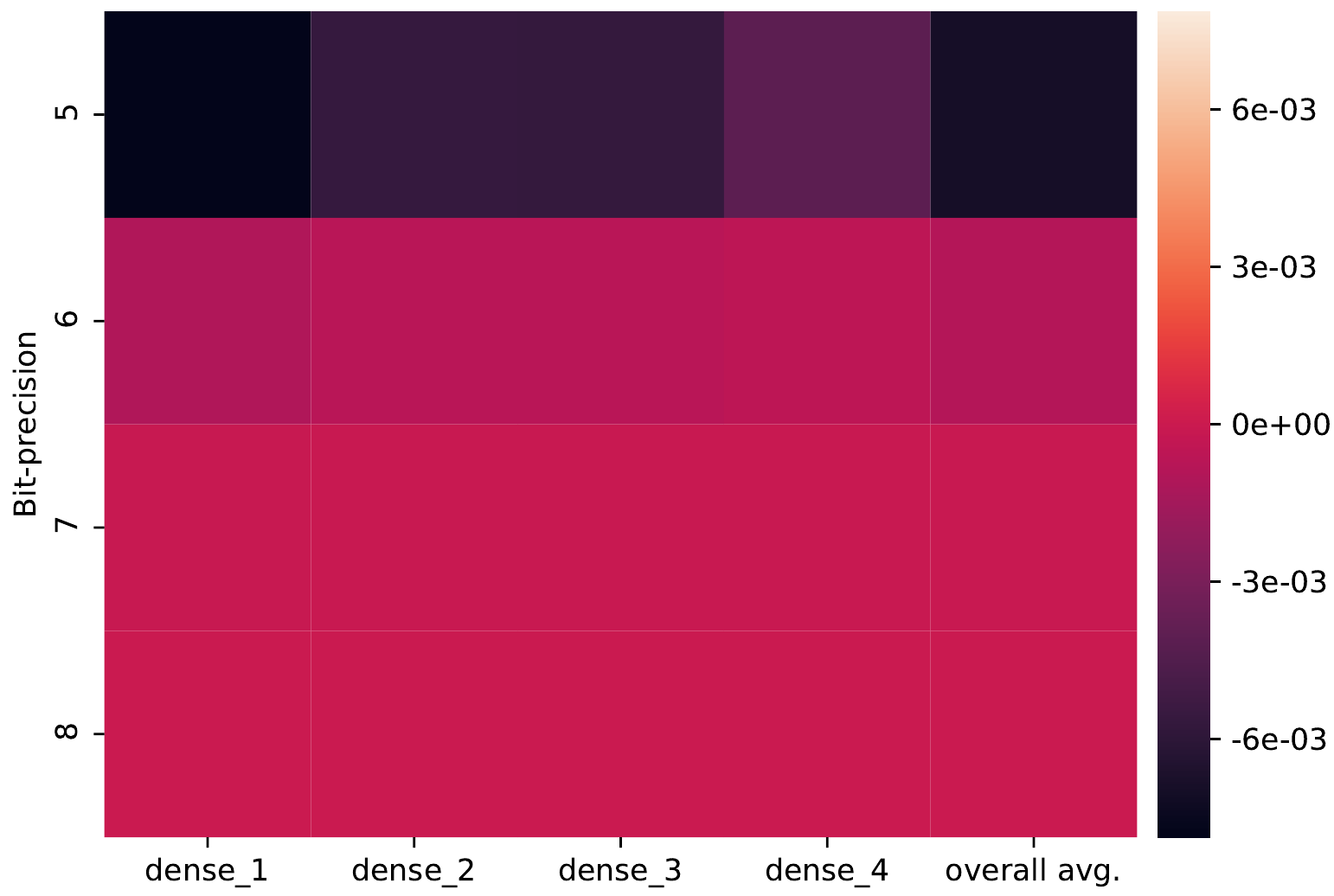}
\end{subfigure}\vskip\baselineskip\vspace{-1mm}
\begin{subfigure}{.075\linewidth}
  \centering
  \caption{}
\end{subfigure}%
\begin{subfigure}{.30\linewidth}
  \centering
  \includegraphics[width=\linewidth]{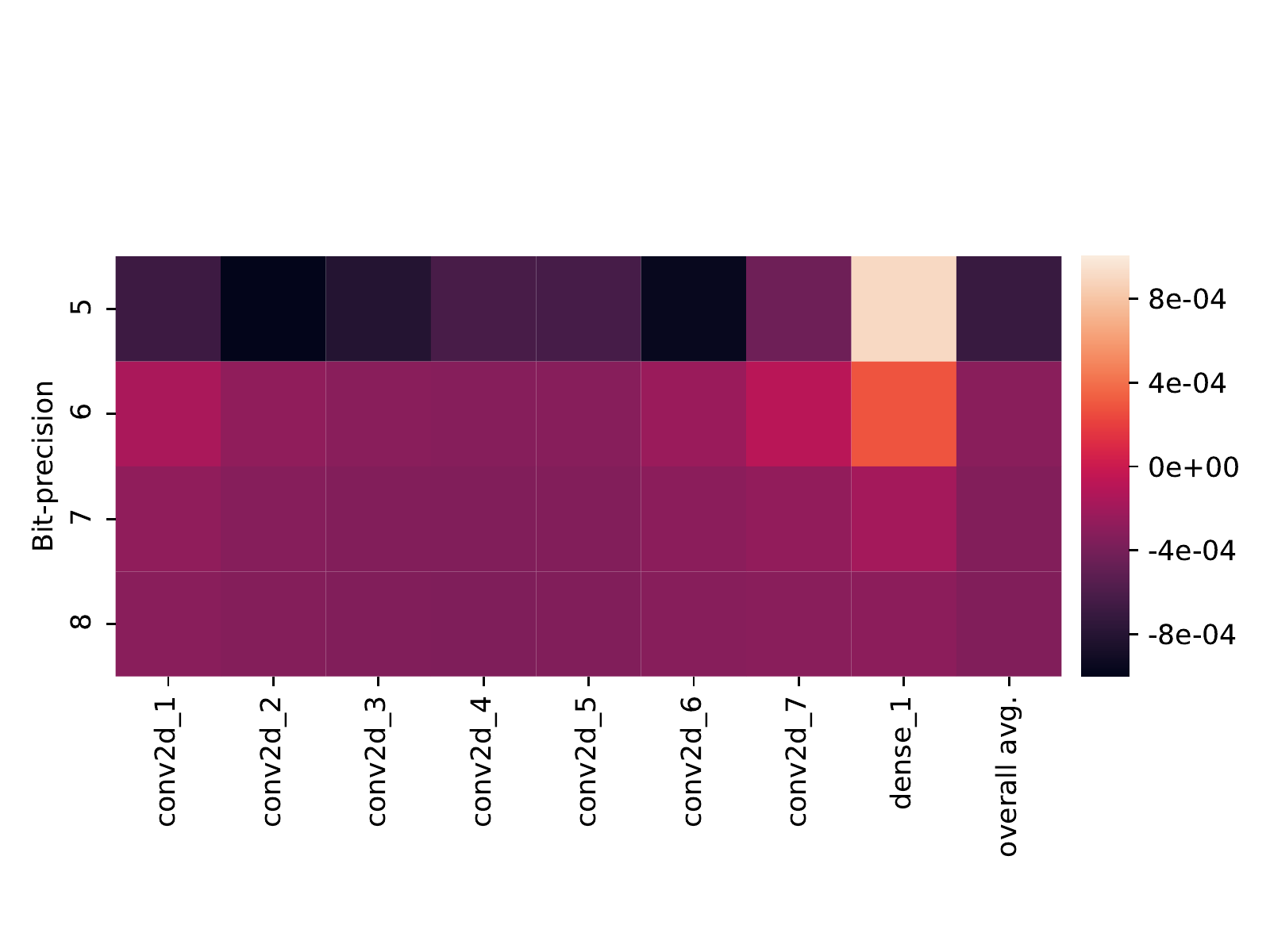}
\end{subfigure}%
\begin{subfigure}{.075\linewidth}
  \centering
  \caption{}
\end{subfigure}%
\begin{subfigure}{.30\linewidth}
  \centering
  \includegraphics[width=\linewidth]{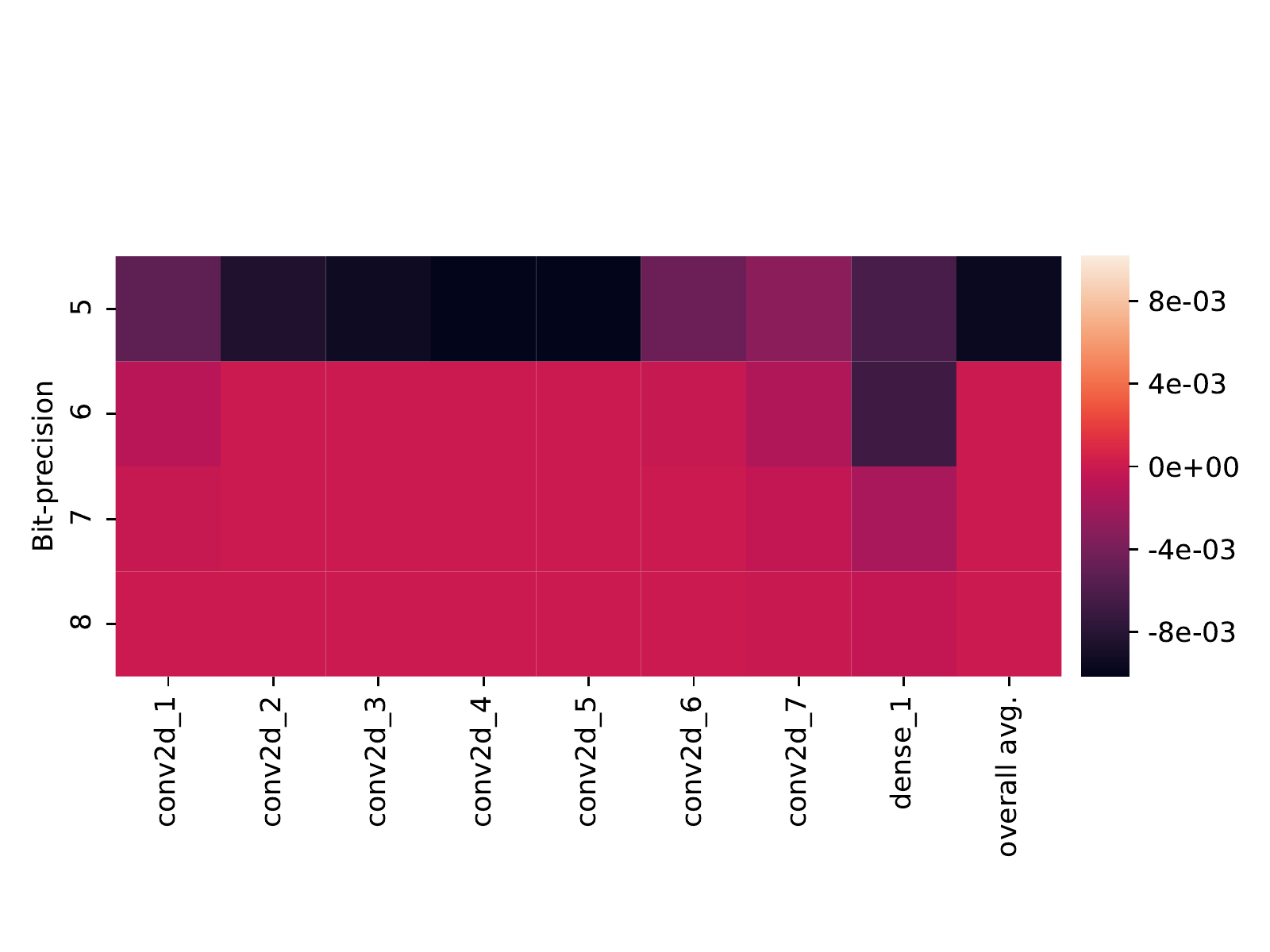}
\end{subfigure}
\caption{Layer-wise delta distortion rate $\Delta(d(R))$ heatmaps compare the precision (rates) of [5..8]-bit numerical formats for representing 32-bit floating point DNN parameters. The average $\Delta(d(R))$ among all weights in a DNN are shown in the final column of each heatmap. (a) $\textup{d(R)}_{posit}-\textup{d(R)}_{fixed}$ for the MNIST task; (b) $\textup{d(R)}_{posit}-\textup{d(R)}_{fixed}$ for the Fashion MNIST task; (c) $\textup{d(R)}_{posit}-\textup{d(R)}_{fixed}$ for the CIFAR-10 task; (d) $\textup{d(R)}_{posit}-\textup{d(R)}_{float}$ for the MNIST task; (e) $\textup{d(R)}_{posit}-\textup{d(R)}_{float}$ for the Fashion MNIST task; (f) $\textup{d(R)}_{posit}-\textup{d(R)}_{float}$ for the CIFAR-10 task.}
\label{fig:mse}
\end{figure*}

\subsection{Exploiting Numerical Formats for DNN Inference}
To evaluate \emph{Cheetah} performance on DNN inference, \textcolor{black} {a feedforward neural network and different convolutional neural networks are} trained on three benchmarks with 32-bit floating point. The specification of these tasks and \textcolor{black} {inference} performance are summarized in Table \ref{table:Spec-32bit-accuracy}. The \textcolor{black} {accuracies of performing DNN inference on these tasks} are presented in Table \ref{table:Accuracy-Result} in the [5..8]-bit precision version of \emph{Cheetah}. The results show that posit with [5..8]-bit precision (mostly $es=1$) outperforms the fixed-point and floating point format\textcolor{black}{s} (mostly $w_e=4$ exponential bits). For instance, the accuracy of performing DNN inference on Fashion-MNIST is improved by \textcolor{black} {$5.14\%$ and $4.17\%$} with 5-bit posits in comparison to 5-bit \textcolor{black} {floating point and fixed-point, respectively. On the CIFAR-10 dataset, these performance gains are further noticeable with 5-bit posits having $28.5\%$ and $31.62\%$ improvements over floating point and fixed-point, respectively.}     
The \textcolor{black} {benefits} of the posit numerical format are intuitively explained by the nonlinear distribution of its values, similar to that of DNN inference parameters. 
This hypothesis is explored empirically by calculating the distortion rate of DNN inference parameters with respect to each numerical format. The distortion rate is described by \eqref{Equ:RD} where $P$ indicates the high-precision parameters and $Quant(P)$ represents the quantized parameters. The results, \textcolor{black} {as shown in Fig. \ref{fig:mse},} validate the hypothesis, especially at 5\textcolor{black}{-bit precision} where the distortion rate of posit is significantly less than that of the other numerical formats.

\begin{equation}\label{Equ:RD}
\begin{split}
d(R) =&~d(P, Quant(P)) = \frac{1}{n} \sum_i^n ||P_i, Quant(P_i)||_2
\end{split}
\end{equation}

\begin{table*}[!ht]
\vspace*{0.5 cm}
\caption{Comparison of different quantization approaches. Accuracy on MNIST (top) \textcolor{black}{and} Fashion-MNIST (bottom) with \{5-8\}-bit precisio\textcolor{black}{n} for posit with $es \in \{0, 1, 2\}$, fixed-point\textcolor{black}{,} and floating point with exponent bit-width $w_e \in \{3, 4\}$.} \label{table:Linear-Quantization}
\vspace*{-1mm}
\centering
\ra{1.2}
\begin{threeparttable}
\resizebox{.99\linewidth}{!}{
\begin{tabular}{@{}ccccccccccccccc@{}} 
 \toprule
    \multirow{2}{*}{Numerical \textcolor{black}{F}ormat} & \multicolumn{4}{c}{Rounding Quantization} && \multicolumn{4}{c}{Linear-Quantization with \textcolor{black}{M}ultiplication} && \multicolumn{4}{c}{Linear-Quantization with \textcolor{black}{S}hift} \\
    \cmidrule{2-5}\cmidrule{7-10}\cmidrule{12-15}
     & 8-bit & 7-bit & 6-bit & 5-bit && 8-bit & 7-bit & 6-bit & 5-bit && 8-bit & 7-bit & 6-bit & 5-bit \\
    \midrule
    Posit ($es=0$) & {98.42}\% & {98.37}\% & {98.30}\% & {91.05}\% && {98.46}\% & {98.48}\% & \textbf{98.46}\% & {98.19}\% && \textbf{98.48}\% & \textbf{98.46}\% & {98.39}\% & {98.28}\% \\
    Posit ($es=1$) & \textbf{98.45}\% & \textbf{98.39}\% & {98.34}\% & \textbf{98.30}\% && \textbf{98.49}\% & {98.47}\% & {98.42}\% & \textbf{98.34}\% && \textbf{98.48}\% & 98.42\% & {98.38}\% & \textbf{98.42}\% \\
    Posit ($es=2$) & {98.44}\% & \textbf{98.39}\% & \textbf{98.37}\% & {98.16}\% && 98.45\% & \textbf{98.49}\% & 98.38\% & 97.96\% && {98.46}\% & 98.41\% & \textbf{98.41}\% & 98.13\%\\
    
    Fixed-point & 98.31\% & 97.95\% &  97.87\% & {97.88}\% && {98.47}\% & {98.32}\% & 98.11\% & 96.41\% && 98.42\% & 98.29\% & {98.16}\% & 97.17\%\\
    
    Floating point & 98.42\% & \textbf{98.39}\% &  98.33\% & 93.91\% && 98.46\% & 98.42\% & 98.36\% & 98.02\% && {98.46}\% & {98.45}\% & {98.38}\% & {98.06}\%\\
    \midrule
    32-bit Floating point & \multicolumn{4}{c}{98.46\%} && \multicolumn{4}{c}{98.46\%} && \multicolumn{4}{c}{98.46\%} \\
    \bottomrule
\end{tabular}
}
\vspace*{-1mm}
\end{threeparttable}
\end{table*}
\begin{table*}[ht!]
\vspace*{0. cm}
\vspace*{-1mm}
\centering
\ra{1.2}
\begin{threeparttable}
\resizebox{.99\linewidth}{!}{
\begin{tabular}{@{}ccccccccccccccc@{}} 
 \toprule
    \multirow{2}{*}{Numerical \textcolor{black}{F}ormat} & \multicolumn{4}{c}{Rounding Quantization} && \multicolumn{4}{c}{Linear-Quantization with \textcolor{black}{M}ultiplication} && \multicolumn{4}{c}{Linear-Quantization with \textcolor{black}{S}hift} \\
    \cmidrule{2-5}\cmidrule{7-10}\cmidrule{12-15}
     & 8-bit & 7-bit & 6-bit & 5-bit && 8-bit & 7-bit & 6-bit & 5-bit && 8-bit & 7-bit & 6-bit & 5-bit \\
    \midrule
    Posit ($es=0$) & {89.57}\% & {89.21}\% & {88.46}\% & {76.87}\% & & \textbf{89.64}\% & \textbf{89.58}\% & \textbf{89.36}\% & {88.17}\% && {89.59}\% & \textbf{89.61}\% & {88.31}\% & {88.10}\% \\
    Posit ($es=1$) & \textbf{89.59}\% & \textbf{89.44}\% & {89.22}\% & \textbf{88.14}\% && {89.58}\% & {89.52}\% & {89.35}\% & \textbf{88.98}\% && {89.58}\% & {89.45}\% & \textbf{89.48}\% & \textbf{89.07}\% \\
    Posit ($es=2$) & {89.56}\% & {89.33}\% & \textbf{89.24}\% & {87.07}\% &&  {89.53}\% & {89.55}\% & {88.98}\% & {87.06}\% && {89.49}\% & {89.52}\% & {89.18}\% & {87.06}\% \\
    
    Fixed-point & {89.16}\% & {87.27}\% & {85.20}\% & {83.97}\% && {89.52}\% & {88.83}\% & {87.46}\% & {76.58}\% && {89.40}\% & {88.93}\% & {87.10}\% & {82.10}\%\\
    
    Floating point & {89.56}\% & {89.36}\% &  {88.92}\% & {83.00}\% && {89.59}\% & {89.45}\% & {89.00}\% & {87.25}\% && \textbf{89.73}\% & {89.32}\% & {88.86}\% & {87.37}\% \\
    \midrule
    32-bit Floating point & \multicolumn{4}{c}{89.51\%} && \multicolumn{4}{c}{89.51\%} &&  \multicolumn{4}{c}{89.51\%} \\
    \bottomrule
\end{tabular}
}
\vspace*{+0.90mm}
\end{threeparttable}
\end{table*}

\subsection{Exploiting Numerical Formats with Quantization Approaches for DNN Inference}
As mentioned before, quantization with rounding has less overhead when compared to the other quantization approaches, but it is not possible to perform DNN inference with 5-bit posits with similar performance of DNN inference \textcolor{black}{as} 32-bit floating point.
To improve performance of DNN inference, the \textcolor{black}{[5..8]-bit posit numerical format is combined with linear quantization approaches and evaluated for a 4-layer feedforward neural network on the MNIST and Fashion-MNIST datasets. The $\alpha_1 \times A_i$ and $\alpha_2 \times W_{ij}$ in \eqref{eq:FeedForwardNetwork-Convolution-Quantized} can be either implemented by constant multiplication or by a shift operation where the $\alpha_1$ and $\alpha_2$ values are approximated by a power of two}.  
The results, \textcolor{black} {as shown in Table \ref{table:Linear-Quantization}}, exhibit that 5-bit low-precision DNN inference achieves similar performance to 32-bit floating point DNN inference on the MNIST data set. Essentially, by deploying this approach, the quantization error produced by the values \textcolor{black}{that} lie outside of posit's dynamic range is zeroed out. The linear quantization approach also plays a key \textcolor{black}{role in reducing} the hardware complexity of posit EMAC\textcolor{black}{s} used for DNN inference. \textcolor{black}{Notably}, the accuracy of DNN inference with posits is significantly enhanced by using \textcolor{black}{the} linear quantization approach in comparison to quantization with rounding. Therefore, the overhead of adding linear quantization is offset by reducing the hardware complexity, i.e. carrying out the posit EMAC operation with $es=0$ instead of $es=1$, which is explained in depth in the next section.

\subsection{Exploiting Posit and Floating Point for DNN Training}
To explore the efficacy of the posit numerical format over the floating point numerical format, a 4-layer feedforward neural network is trained with each number system on the MNIST and Fashion-MNIST datasets. The results indicate that the posit numerical format has a slightly better accuracy in comparison to the floating point number system, \textcolor{black}{as shown in Table \ref{tab:train_acc}}. \textcolor{black} {16-bit posits outperform 16-bit floats in terms of accuracy. Although \emph{Cheetah} is evaluated on small datasets, there are two advantages compared to \cite{IBM8,mellempudi2019mixed}. Mellempudi \textit{et al.} \cite{mellempudi2019mixed} use 32-bit numbers for accumulation to reduce the hardware cost of stochastic rounding. Wang \textit{et al.} \cite{IBM8} reduce the accumulation bit-precision to 16 by using stochastic rounding. However, in this paper, we show the potential of using 16-bit posits for \textit{all DNN parameters} with a simple and hardware-friendly round-to-nearest algorithm and show less than 1\% accuracy degradation without exhaustively analyzing DNN training parameters.}

\begin{table}[H]
\caption{Average accuracy over 10 \textcolor{black}{independent} runs \textcolor{black}{on the test set of the respective dataset. Networks are trained using only the specified numerical format}.}\label{tab:train_acc}
\centering
\ra{1.2}
\begin{threeparttable}
\begin{tabular}{@{}lcc@{}}
    \toprule
    Task          & Format   & Accuracy \\
    \midrule
    \multirow{4}{*}{MNIST}         & Posit-32 & \textbf{98.131\%} \\
                                   & Float-32 &         98.087\%  \\
                                   & Posit-16 &         96.535\%  \\
                                   & Float-16 &         90.646\%  \\[+1ex]
    \multirow{4}{*}{Fashion MNIST} & Posit-32 & \textbf{89.263\%} \\
                                   & Float-32 &         89.105\%  \\
                                   & Posit-16 &         87.400\%  \\
                                   & Float-16 &         81.725\%  \\
    \bottomrule
\end{tabular}
\end{threeparttable}
\end{table}


\subsection{EMAC Soft-Core FPGA Implementation}
\begin{figure*}
    \centering
    \includegraphics[width=.75\linewidth]{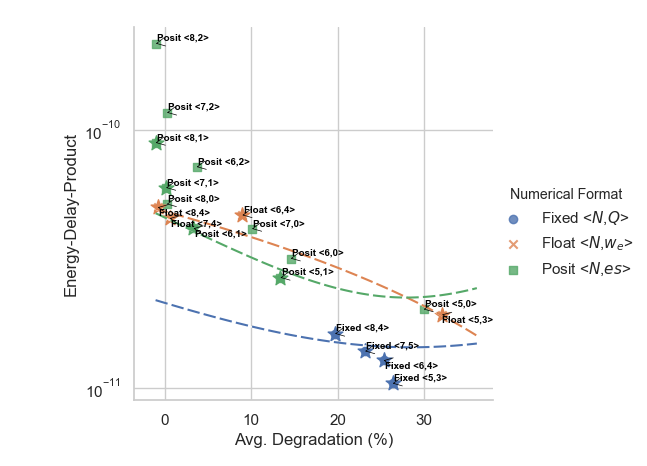}
    \caption{The average accuracy degradation from 32-bit floating point across the two classification tasks vs. the energy-delay-product of the respective EMAC with rounding quantization. \textcolor{black} {Each $<$x, y$>$ pair indicates the number of bits and corresponding parameter bit-width, as indicated in the legend.} A star ($\star$) denotes the lowest accuracy degradation for a numerical format and bit-width.}
    \label{fig:acc_EDP}
\end{figure*}

\begin{figure*}
    \centering
    \includegraphics[width=.75\linewidth]{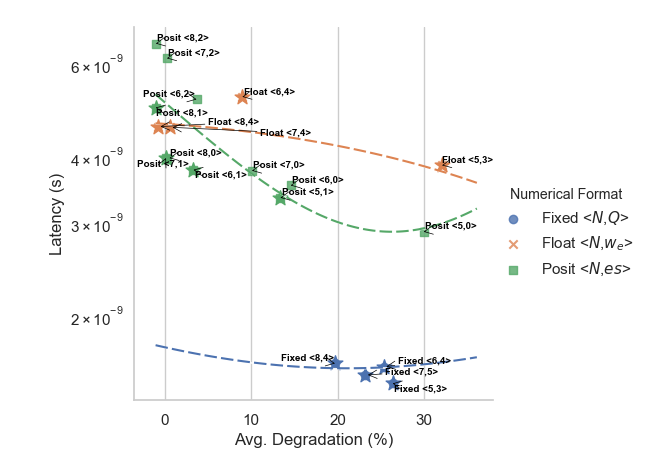}
    \caption{The average accuracy degradation from 32-bit floating point across the two classification tasks vs. the latency of the respective EMAC with rounding quantization. \textcolor{black} {Each $<$x, y$>$ pair indicates the number of bits and corresponding parameter bit-width, as indicated in the legend.} A star ($\star$) denotes the lowest accuracy degradation for a numerical format and bit-width.}
    \label{fig:acc_Latancy}
\end{figure*}

\begin{figure*}
    \centering
    \includegraphics[width=.80\linewidth]{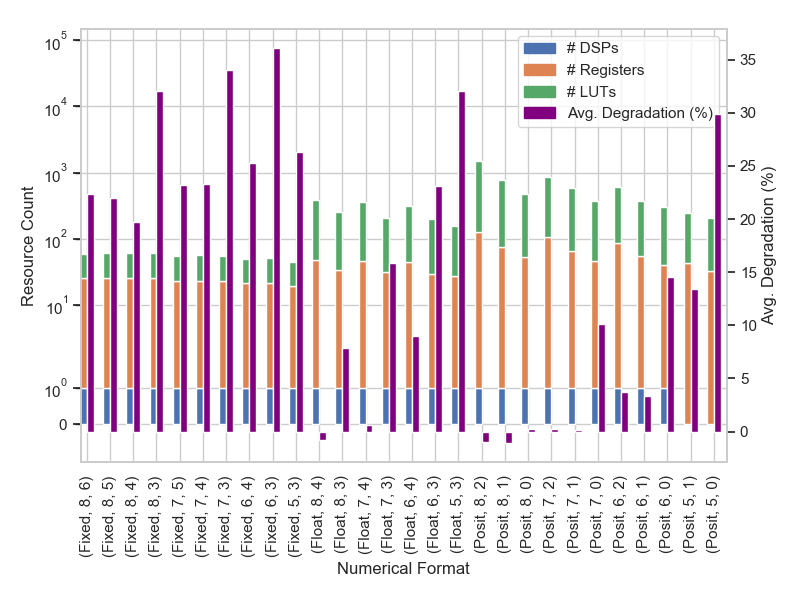}
    \caption{The average accuracy degradation from 32-bit floating point across the two classification tasks vs. the cost of the respective EMAC with rounding quantization. \textcolor{black} {Each $<$x, y$>$ pair indicates the number of bits and corresponding parameter bit-width (fractional bits or exponent), as labeled along the x-axis.}}
    \label{fig:acc_cost}
\end{figure*}

\begin{figure*}
    \centering
    \includegraphics[width=.70\linewidth]{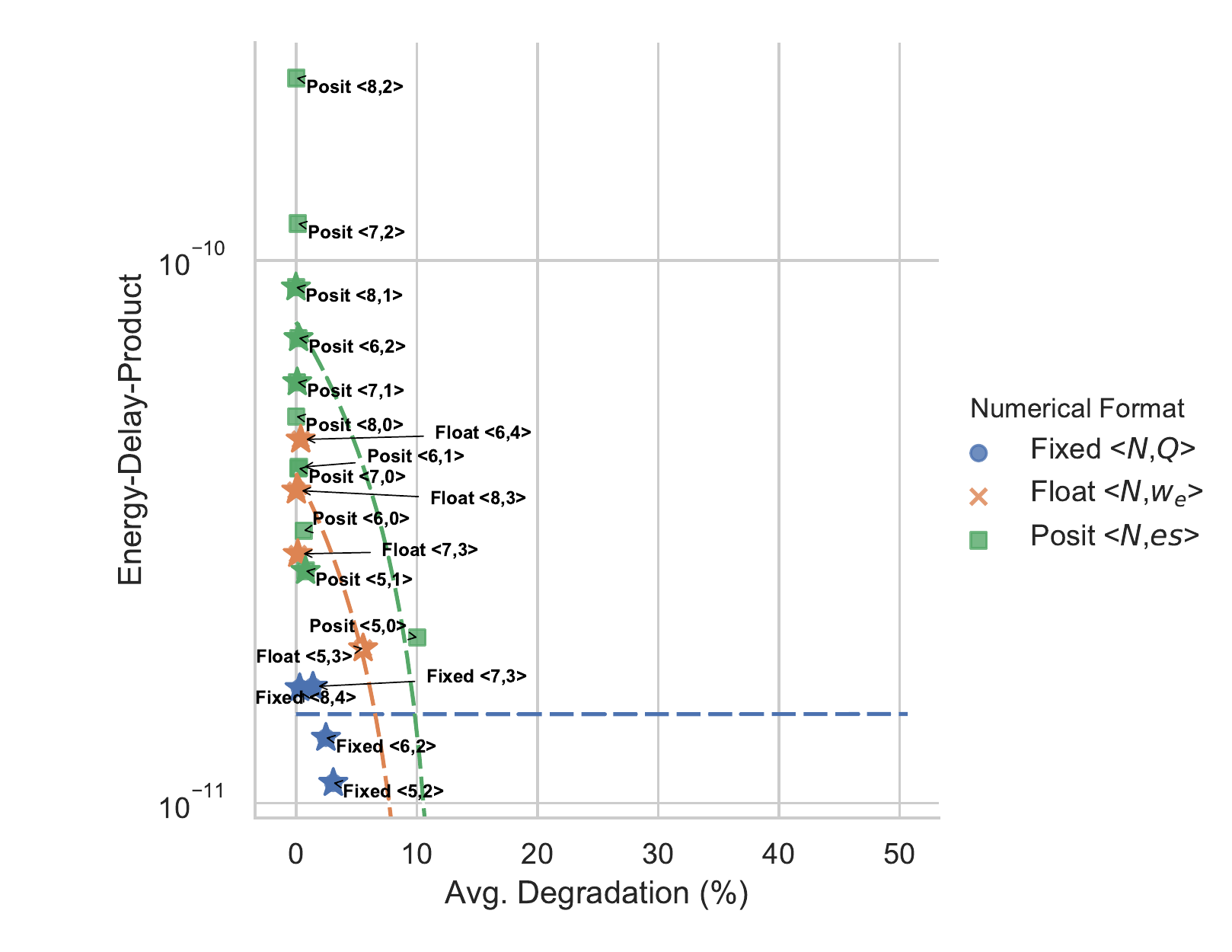}
    \caption{The average accuracy degradation from 32-bit floating point across the two classification tasks vs. the energy-delay-product of the respective EMAC with linear quantization. \textcolor{black} {Each $<$x, y$>$ pair indicates the number of bits and corresponding parameter bit-width, as indicated in the legend.} A star ($\star$) denotes the lowest accuracy degradation for a numerical format and bit-width.}
    \label{fig:acc_EDP-Linear}
\end{figure*}

\begin{figure*}
    \centering
    \includegraphics[width=.70\linewidth]{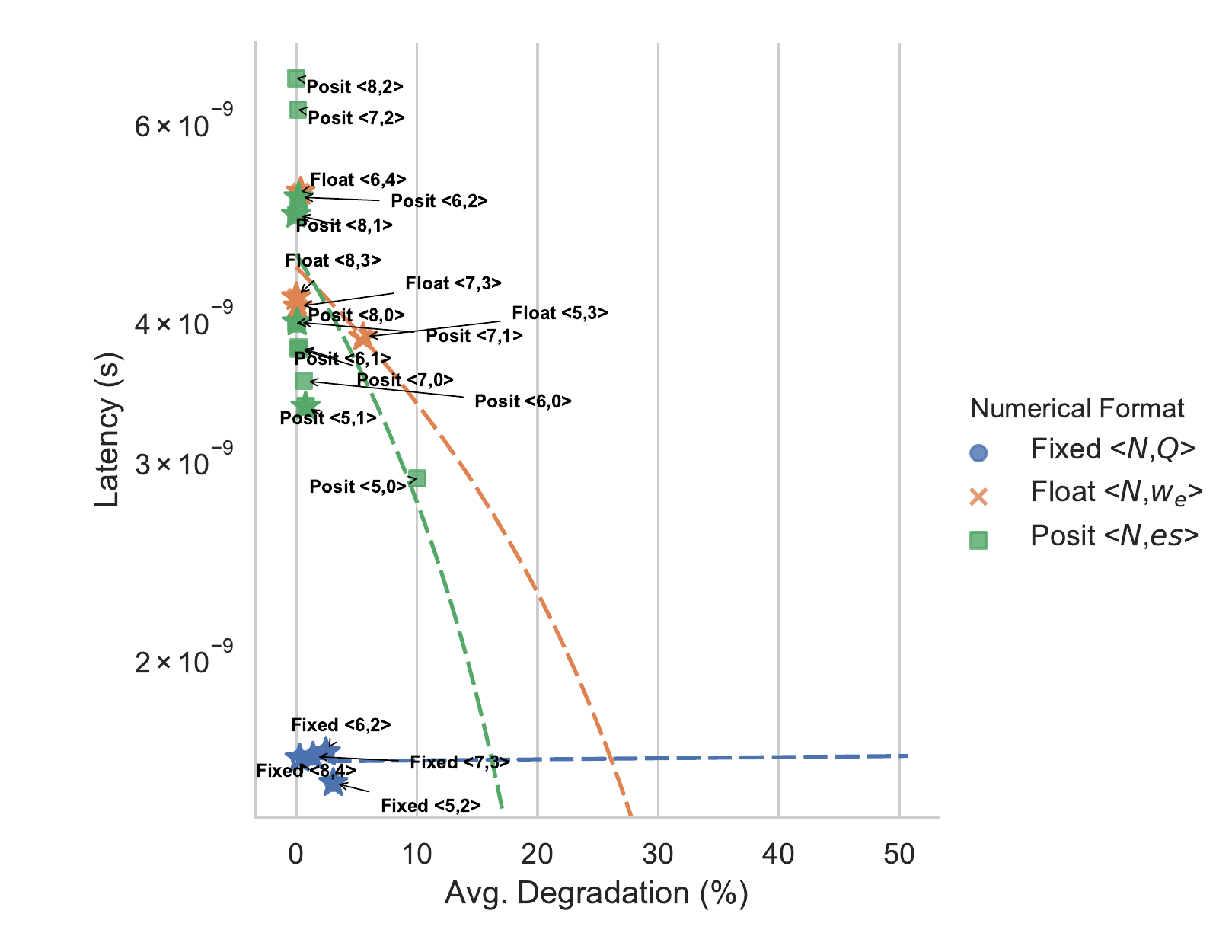}
    \caption{The average accuracy degradation from 32-bit floating point across the two classification tasks vs. the latency of the respective EMAC with linear quantization. \textcolor{black} {Each $<$x, y$>$ pair indicates the number of bits and corresponding parameter bit-width, as indicated in the legend.} A star ($\star$) denotes the lowest accuracy degradation for a numerical format and bit-width.}
    \label{fig:acc_latency-Linear}
\end{figure*}

To \textcolor{black}{show} the effectiveness of the posit numerical format over floating point and fixed-point,
we evaluate the trade-off between the energy-delay-product and \textcolor{black} {latency} of the EMAC operation \textcolor{black} {vs.} average accuracy degradation from 32-bit floating point per bit-width across the \textcolor{black} {three datasets (two for the linear-quantization experiment) with} the \emph{Cheetah} framework\textcolor{black}{,} as shown in \textcolor{black} {Figs. \ref{fig:acc_EDP},  \ref{fig:acc_Latancy}, \ref{fig:acc_cost}, \ref{fig:acc_EDP-Linear}, and \ref{fig:acc_latency-Linear}}. The energy-delay-product%
\textcolor{black}{, a combined measure of the latency and resource cost of the EMAC operation,} coupled with quantization with rounding \cite{carmichael2019positron} and the EMAC operation coupled with linear quantization are selected for all numerical formats and measured on a Virtex-7 FPGA (xc7vx485t-2ffg1761c) \textcolor{black}{with} \textcolor{black}{synthesis} through Vivado 2017.2. Note that the average accuracy degradation per bit-width is computed \textcolor{black}{using the} accuracy results \textcolor{black}{in} Table \ref{table:Linear-Quantization}.

The results\textcolor{black}{,} as shown by Fig. \ref{fig:acc_EDP}\textcolor{black}{,} \textcolor{black}{indicate} tha\textcolor{black}{t} posit coupled with rounding quantization \textcolor{black}{achieves} up to \textcolor{black} {23\% average accuracy improvement over fixed-point.} However, this accuracy enhancement is gained at the cost of \textcolor{black}{a} $0.41 \times 10^{-10}$ increase in energy-delay-product to implement the EMAC unit. Posit also consistently shows better performance, especially at 5-bit compared to \textcolor{black}{the} floating point number system at a \textcolor{black} {comparable energy-delay-product}. \textcolor{black} {The posit EMAC operation achieves lower latencies, as shown in Fig. \ref{fig:acc_Latancy}, due to a lack of subnormal detection and other exception cases, but exhibits resource-hungry encoding and decoding due to the variable-length regime of the posit numerical format, as shown in Fig. \ref{fig:acc_cost}.} Overall, the 6-bit posit shows \textcolor{black}{the} best trade-off between energy-delay-product and average accuracy degradation from 32-bit floating point on the two benchmarks (when analyzed across \textcolor{black}{the} [5..8]-bit range). \textcolor{black}{Looking at the posit numerical format in terms of classification performance and EMAC energy-delay-product, posits with $es=1$ provide a better trade-off compared to posits with $es \in \{0, 2\}$. At [5..7]-bit precision, the average performance of DNN inference with $es=1$ among the three datasets is 2\% and 4\% better than with $es=2$ and $es=0$, respectively. These accuracy benefits are coupled with 2.1$\times$ less energy-delay-product and 1.4$\times$ more energy-delay-product in comparison to $es=2$ and $es=0$, respectively. These results are measured when the rounding quantization is used. L}inear quantization with the shift operation requires similar hardware overhead across all \textcolor{black}{of} the numerical formats, as shown in Figs. \ref{fig:acc_EDP-Linear} and \ref{fig:acc_latency-Linear}. However, the accuracy of performing DNN inference with linear quantization \textcolor{black}{with posits} ($es=0$) is similar to the accuracy \textcolor{black}{when $es=1$}. Therefore, it is possible to use EMAC\textcolor{black}{s with} $es=0$ instead of $es=1$ and thereby achieve 18\% energy-delay-product savings.




A summary of previous studies that propose low-precision frameworks are shown in Table \ref{tab:compare}. Several research groups have explored the efficacy of floats and fixed-point on the performance and hardware complexity of DNNs with multiple image classification tasks \cite{Courbariaux14,hashemi2017,gysel2018,IBM8,flexpoint2017,micikevicius2017mixed}. However, none of these works analyze the appropriateness of the posit numerical format for both DNN training and inference. Additionally, current work does not offer insight on the impact of the quantization approach vs. numerical format on both accuracy and hardware complexity, as investigated in this paper.

\begin{table*}
   \caption{High-level summary of \emph{Cheetah} and other low-precision frameworks. All datasets are image classification tasks. WI BC: Wisconsin Breast Cancer; FMNIST: Fashion MNIST; FP: floating point; FX: fixed-point; PS: posit; SW: software; HW: hardware.}
   \label{tab:compare}
   \ra{1.3}
   \centering
   \resizebox{\linewidth}{!}{%
   \begin{tabular}{@{}cccccccc@{}}
     \toprule
     & Courbariaux \textit{et al.} \cite{courbariaux2014training}& Gysel \textit{et al.} \cite{gysel2018} & Hashemi \textit{et al.} \cite{hashemi2017} & Carmichael \textit{et al.} \cite{carmichael2019positron} & Wang  \textit{et al.} \cite{IBM8} & Johnson \textit{et al.} \cite{johnson2018rethinking} & This Work \\
     \midrule
     \multirow{2}{*}{Dataset} & \multirow{1}{*}{MNIST, \textcolor{black}{CIFAR}-10,} & \multirow{2}{*}{ImageNet} & \multirow{1}{*}{MNIST, \textcolor{black}{CIFAR}-10,} & WI BC, Iris, Mushroom & \multirow{2}{*}{ImageNet} & \multirow{2}{*}{ImageNet} & MNIST, FMNIST \\[-1ex]
     & SVHN && SVHN & MNIST, FMNIST &&& \textcolor{black}{CIFAR}-10 \\
     \multirow{2}{*}{Numerical Format} & FP, FX, & FP, FX, & FP, FX & FP, FX & \multirow{2}{*}{FP} & FX, FP & FX, FP \\[-1ex]
     & \textcolor{black}{B}FP & \textcolor{black}{B}FP & Binary & PS & & PS & PS \\
     Bit-precision & 12 & 8 & All & [5..8] & All & 8 & [5..8] \\ 
     Utility & Training & Inference & Inference & Inference & Training & Inference & Inference \& Training \\
     Inference Quantization & - & Rounding & Rounding & Rounding & - & Log & Rounding \& \textcolor{black}{L}inear \\
     Implementation & SW & SW \& HW & SW \& HW & SW \& HW & SW \& HW & SW \& HW & SW \& HW \\
     DNN library & Theano & Caffe & Caffe & Keras/TensorFlow & Home \textcolor{black}{S}uite & PyTorch & Keras/TensorFlow \\
     \multirow{1}{*}{Device} & \multirow{1}{*}{-} & \multirow{1}{*}{ASIC} & ASIC & Virtex-7 FPGA & \multirow{1}{*}{ASIC} & \multirow{1}{*}{ASIC} & Virtex-7 FPGA  \\ 
     Technology Node & - & 65 nm & 65 nm & 28 nm & 14 nm & 28 nm & 28 nm\\
     \bottomrule
   \end{tabular}
   }
   \label{tab:OthersWork}
 \end{table*}

\section{Conclusions}
A low-precision DNN framework, \emph{Cheetah}, for edge devices is proposed in this work. We explored the \textcolor{black}{capacity} of various numerical formats\textcolor{black}{,} including floating point, fixed-point and posit\textcolor{black}{, for} both DNN training and inference. We \textcolor{black}{show} that the recent posit numerical format has high efficacy for DNN training at \textcolor{black}{\{16, 32\}}-bit precision and inference at $\leq$8-bi\textcolor{black}{t} precision. Moreover, we show that it is possible to achieve better performance and reduce energy consumption by using linear quantization \textcolor{black}{with} the posit numerical format. The success of low-precision posit\textcolor{black}{s in} reduc\textcolor{black}{ing} DNN hardware complexity with negligible accuracy degradatio\textcolor{black}{n} motivates us to evaluate ultra\textcolor{black}{-}low precision training \textcolor{black}{in} future work.

\ifCLASSOPTIONcaptionsoff
  \newpage
\fi

\bibliographystyle{IEEEtran}
\bibliography{Journal.bib}









\end{document}